\documentclass[lettersize,journal]{IEEEtran}
\usepackage{amsmath,amsfonts}
\usepackage{algorithmic}
\usepackage{algorithm}
\usepackage{array}
\usepackage{textcomp}
\usepackage{stfloats}
\usepackage{verbatim}
\usepackage{graphics} % for pdf, bitmapped graphics files
\usepackage{epsfig} % for postscript graphics files
\usepackage{amsmath} % assumes amsmath package installed
\usepackage{amssymb}  % assumes amsmath package installed
\usepackage{times} % assumes new font selection scheme installed
\usepackage{cite} % 导入引用的包，能够使用\cite
\usepackage{graphicx}
\usepackage{subfigure} %子图
\usepackage{picinpar}
\usepackage{url}
\usepackage{flushend}
\usepackage{colortbl}
\usepackage{soul}
\soulregister{\cite}7 % 注册\cite命令
\soulregister{\citep}7 % 注册\citep命令
\soulregister{\citet}7 % 注册\citet命令
\soulregister{\ref}7 % 注册\ref命令
\soulregister{\pageref}7 % 注册\pageref命令

\usepackage{multirow}
\usepackage{pifont}
\usepackage{alltt}
\usepackage{hyperref} %点击链接 
\usepackage{float}%需要把图片放在当前位置，不容改变，可以用float宏包的[H]选项
\usepackage{enumerate}
\usepackage{siunitx}
\usepackage{breakurl}
\usepackage{epstopdf}
\usepackage{pbox}
\usepackage{booktabs}
\usepackage{makecell} %表格换行
\usepackage{threeparttable} %加脚注
\usepackage{balance}%最后一页 参考文献不平衡
\usepackage[justification=centering,labelsep=period]{caption} %全局图片标题居中
\usepackage{color} %颜色
\usepackage{gensymb} %度
\hypersetup{colorlinks=true, linkcolor=black} %修改链接颜色

\usepackage{adjustbox}%用于highlight

\usepackage{bm}
\usepackage{bbm} %小写字母的\mathbb 
\usepackage{geometry}
\title{\LARGE \bf
    DEIO: Deep Event Inertial Odometry 
}

\author{Weipeng~Guan$^{*}$,   Fuling~Lin$^{*}$,   Peiyu~Chen,   Peng~Lu  % <-this % stops a space
    \thanks{
        *Equal contribution; 
        The authors are with the Adaptive Robotic Controls Lab (ArcLab), Department of Mechanical Engineering, The University of Hong Kong, Hong Kong SAR, China.
        (corresponding author: lupeng@hku.hk). 
        }%
    \thanks{
        This work was supported by General Research Fund under Grant 17204222, and in part by the Seed Fund for Collaborative Research and General Funding Scheme-HKU-TCL Joint Research Center for Artificial Intelligence.
        }%
}

\begin{document}   %正式开始这个文档
\maketitle  %不要加页面，避免编译不过
\thispagestyle{headings} 
\pagestyle{headings}

%%%%%%%%%%%%%%%%%%%%%%%%%%%%%%%%%%%%%%%%%%%%%%%%%%%%%%%%%%%%%%%%%%%%%%%%%%%%%%%%%%%%%%%%%%%%%%%%%%%%%%%%%%%%
\begin{abstract}
Event cameras show great potential for visual odometry (VO) in handling challenging situations, such as fast motion and high dynamic range. 
Despite this promise, the sparse and motion-dependent characteristics of event data continue to limit the performance of feature-based or direct-based data association methods in practical applications. 
To address these limitations, we propose Deep Event Inertial Odometry (DEIO), the first monocular learning-based event-inertial framework, which combines a learning-based method with traditional nonlinear graph-based optimization. 
Specifically, an event-based recurrent network is adopted to provide accurate and sparse associations of event patches over time.
DEIO further integrates it with the IMU to recover up-to-scale pose and provide robust state estimation.
The Hessian information derived from the learned differentiable bundle adjustment (DBA) is utilized to optimize the co-visibility factor graph, which tightly incorporates event patch correspondences and IMU pre-integration within a keyframe-based sliding window.
Comprehensive validations demonstrate that DEIO achieves superior performance on \textit{10} challenging public benchmarks compared with more than \textit{20} state-of-the-art methods. 
\end{abstract}

%%%%%%%%%%%%%%%%%%%%%%%%%%%%%%%%%%%%%%%%%%%%%%%%%%%%%%%%%%%%%%%%%%%%%%%%%%%%%%%%%%%%%%%%%%%%%%%%%%%%%%%%%%%%
% \begin{IEEEkeywords} 
% Event-based Vision, 6-DoF Pose Tracking, Event Camera, SLAM, Deep Learning, Robotics.
% \end{IEEEkeywords} 

%%%%%%%%%%%%%%%%%%%%%%%%%%%%%%%%%%%%%%%%%%%%%%%%%%%%%%%%%%%%%%%%%%%%%%%%%%%%%%%%
% \vspace{-1.0em}%调整与上文的距离
\section*{MULTIMEDIA MATERIAL}  %不带序号
\label{MULTIMEDIA MATERIAL}
{\footnotesize \textbf{Website}: \url{https://kwanwaipang.github.io/DEIO/}.}

{\footnotesize \textbf{Codes}: \url{https://github.com/arclab-hku/DEIO}.}

%%%%%%%%%%%%%%%%%%%%%%%%%%%%%%%%%%%%%%%%%%%%%%%%%%%%%%%%%%%%%%%%%%%%%%%%%%%%%%%%%%%%%%%%%%%%%%%%%%%%%%%%%%%%
\section{INTRODUCTION}
\label{section: INTRODUCTION}

%%%%%%%%%%%%%%%%%%%%%%%%%%%%%%%%%%%%%%%%%%%%%%%%%%%%%%%%%%%%%%%%%%%%%%%%%%%%%%%%%%%%%%%%%%%%%%%%%%%%%%%%%%%%
\subsection{Motivation}

\IEEEPARstart{A}{chieving} reliable and accurate visual odometry (VO) under adverse conditions remains challenging, particularly when employing image-based solutions (RGB or RGB-D cameras).
Event cameras are motion-activated sensors that only capture pixel-wise intensity changes with microsecond precision and report them as an asynchronous stream instead of the whole scene as an intensity image with a fixed frame rate.
Because of their remarkable properties, such as high temporal resolutions, high dynamic range (HDR), and no motion blur, event cameras have the potential to enable high-quality perception in extreme lighting conditions and high-speed motion scenarios that are currently not accessible to standard cameras.

Despite such promises, integrating event cameras into VO systems presents significant challenges. 
This is primarily due to the sparse, irregular, and asynchronous nature of event data, which conveys limited information and contains inherent noise.
Besides, event cameras face difficulty in capturing visual information when motion is parallel to edges or in a static state.
Moreover, due to the motion-dependent characteristic, both feature-based and direct-based methods easily fail in incomplete observation or sudden variation of the event camera.
Therefore, current purely event-based VO systems~\cite{GWPHKU:EVO,GWPHKU:ESVO} generally lack the robustness requirement for real-world applications. 
Using additional sensors tends to achieve better performance since different modalities can complement each other, such as image frames~\cite{GWPHKU:Ultimate-SLAM, EDS, GWPHKU:PL-EVIO}, depth sensors~\cite{zuo2024cross, zuo2022devo}, or even LiDAR~\cite{FAST-LIEO}.
However, these combinations might limit the application of event cameras in real-world applications due to additional computational costs and more complicated sensor calibration requirements~\cite{GWPHKU:ECMD}.
Additionally, this introduces new bottlenecks.
For instance, relying on image frames would make the system susceptible to motion blur or HDR limitations. 
Recent studies have introduced learning-based approaches~\cite{DEVO,rampvo} as promising solutions for event-based VO, addressing the previously mentioned limitations by employing neural networks to establish robust associations.
These methods, trained exclusively on synthetic event data, demonstrate good generalization to real-world benchmarks and can even outperform some traditional systems in accuracy.
However, it is worth noting that visual/event-only systems have inherent limitations, making them vulnerable to low-textured environments or suffering from scale ambiguity.
To mitigate visual degradations, a practical and promising strategy is to incorporate the Inertial Measurement Unit (IMU), which is low-cost and readily available in most event cameras.
Furthermore, in the minimal configuration of a monocular setup, the IMU can be leveraged to recover the metric scale and enhance the accuracy and robustness of the VO system.
Nevertheless, efficiently integrating learning-based event data association with IMU measurements remains an open problem.

%%%%%%%%%%%%%%%%%%%%%%%%%%%%%%%%%%%%%%%%%%%%%%%%%%%%%%%%%%%%%%%%%%%%%%%%%%%%%%%%%%%%%%%%%%%%%%%%%%%%%%%%%%%%
\subsection{Contributions}
In this work, we propose Deep Event Inertial Odometry (DEIO), the first deep learning-based event-inertial odometry framework.
It is developed based on a learning-optimization framework that leverages neural networks to predict event correspondences and tightly integrates IMU measurements to enhance the robustness of the odometry.
More specifically, our framework decouples network training from IMU integration and operates in two phases: offline training and online optimization.
During the training phase, an event-based recurrent network learns to provide robust data associations of sparse event patches.
At runtime, IMU measurements are tightly integrated with the trained network within a factor graph optimization framework to achieve robust 6-DoF pose tracking.
This design enables us to train the network using more accessible training datasets without requiring IMU data, while still benefiting from IMU measurements during runtime through online optimization.
The event-based recurrent network is built on the patch-based structure~\cite{DPVO}.
The selected event patches are processed using the recurrent optical flow network and the differentiable bundle adjustment (DBA) layer to establish high-confidence data associations for the consecutive event stream.
The Hessian information, derived from the event-based DBA layer, is fed into an event patch-based co-visibility factor graph, facilitating its seamless fusion with the IMU pre-integration via a well-designed keyframe-based sliding window optimization framework.
Our main contributions are summarized as follows:

\begin{enumerate}

\item
We propose a learning-optimization framework that seamlessly integrates the power of deep learning with the efficiency of factor graph optimization.
To the best of our knowledge, this is the first event-inertial odometry framework that employs deep learning for event data association and graph optimization for pose estimation.

\item 
An event-based co-visibility factor graph optimization is proposed to tightly integrate event patch correspondences by deriving Hessian information from the DBA along with IMU pre-integration.

\item 
Extensive experiments on 10 challenging event-based real-world benchmarks demonstrate the superior performance of our DEIO compared with over 20 state-of-the-art methods.

\item 
Codes and the preprocessed event datasets are released to facilitate further research in learning-based event pose tracking.

\end{enumerate}

%%%%%%%%%%%%%%%%%%%%%%%%%%%%%%%%%%%%%%%%%%%%%%%%%%%%%%%%%%%%%%%%%%%%%%%%%%%%%%%%%%%%%%%%%%%%%%%%%%%%%%%%%%%%
\section{Related Work}
\label{section: Related Work}

%%%%%%%%%%%%%%%%%%%%%%%%%%%%%%%%%%%%%%%%%%%%%%%%%%%%%%%%%%%%%%%%%%%%%%%%%%%%%%%%%%%%%%%%%%%%%%%%%%%%%%%%%%%%
\subsection{Traditional Approaches for Event-based VO}
\label{section: Event-based VO/VIO/SLAM}
The first purely event-based VO is presented in~\cite{GWPHKU:kim2016real}, which performed real-time event-based SLAM through three decoupled probabilistic filters that jointly estimate the 6-DoF camera pose, 3D map of the scene, and image intensity. 
EVO~\cite{GWPHKU:EVO} presents a geometric 2D-3D model alignment, utilizing the 2D image-like event representation and the 3D map reconstructed from EMVS\cite{GWPHKU:EMVS}. 
ESVO~\cite{GWPHKU:ESVO} is a stereo event-based VO method to estimate the ego-motion through the 2D-3D edge registration on the time surface. 
EDS~\cite{EDS} proposes an event-image alignment algorithm that minimizes photometric errors between the brightness change from events and the image gradients, enabling 6-DOF monocular VO.
In addition, some approaches rely on a pre-built photometric 3D map~\cite{gallego2017event} or restrict the motion type to rotation-only~\cite{CMax-SLAM}.
However, most of them are prone to failure in challenging scenarios (such as fast motion).

To enhance the robustness of purely event-based VO, existing event-based SLAM methods have demonstrated good performance by incorporating additional sensors.
Notably, event-inertial integration is a widely used approach to address the limitations of event-only SLAM, which provides scale awareness and continuity of estimation with minimal setup requirements.
Zhu et al.~\cite{GWPHKU:Event-based-visual-inertial-odometry} propose the first event-inertial odometry (EIO) method, which fuses events with IMU through the Extended Kalman Filter.
Rebecq et al.~\cite{GWPHKU:ETH-EVIO} propose an optimization-based EIO that detects and tracks features in the edge image, generated from motion-compensated event streams, through traditional image-based feature detection and tracking. 
The tracked features are then combined with IMU measurements via keyframe-based nonlinear optimization.
Another monocular EIO is presented in~\cite{GWPHKU:MyEVIO}, which employs the event-corner features with IMU measurement to deliver real-time and accurate 6-DoF state estimation. 
Chamorro et al.~\cite{chamorro2023event} integrate the event-based line feature with the IMU to achieve high ultra-fast pose estimation.
Ultimate-SLAM~\cite{GWPHKU:Ultimate-SLAM} and PL-EVIO~\cite{GWPHKU:PL-EVIO} investigate the complementary nature of events and images to present an event-image-IMU odometry (EVIO).
EKLT-VIO~\cite{EKLT-VIO} combines the event-based and image-based feature tracker~\cite{GWPHKU:EKLT} as the front end with a filter-based back end to perform the pose estimation for Mars-like sequences.
ESVIO~\cite{ESVIO} proposes the first stereo EIO and EVIO framework to estimate states through temporally and spatially event-corner feature association.
ESVIO\_AA~\cite{ESVO_IMU} extends ESVO~\cite{GWPHKU:ESVO} and presents a direct method for stereo event cameras with an IMU-aided solution.
EVI-SAM~\cite{GWPHKU:EVI-SAM} introduces the first event-based hybrid pose tracking framework, merging feature-based and direct-based methods. 
Additionally, it consistently delivers excellent mapping results in a variety of challenging scenarios, demonstrating performance comparable to learning-based and NeRF-based dense mapping methods.
Despite such remarkable progress, current event-based VO systems still fall short of the robustness required for certain real-world applications, and their performance has hit a bottleneck.
This is caused by the sparse and motion-dependent nature of event data, traditional feature-based or direct-based data association is hardly established in complex scenarios, which motivates the exploration of deep learning approaches.

%%%%%%%%%%%%%%%%%%%%%%%%%%%%%%%%%%%%%%%%%%%%%%%%%%%%%%%%%%%%%%%%%%%%%%%%%%%%%%%%%%%%%%%%%%%%%%%%%%%%%%%%%%%%
\subsection{Learning-based Approaches for Event-based VO}

Zhu et al.~\cite{zhu2019unsupervised} employ a convolutional neural network (CNN) with an unsupervised encoder-decoder architecture to predict pose, flow, and semi-dense depth.
Hidalgo-Carri{\'o} et al.~\cite{hidalgo2020learning} estimate dense depth maps from a monocular event camera through recurrent CNN architecture.
Gehrig et al.~\cite{gehrig2021combining} apply a recurrent neural network (RNN), which maintains an internal state that is updated through asynchronous events or irregular images and can be queried for dense depth estimation at any timestamp.
Ahmed et al.~\cite{ahmed2021deep} develop a deep event stereo network that reconstructs spatial image features from embedded event data and leverages the event features using the reconstructed image features to compute dense disparity maps.
However, most of these methods require knowledge of camera motion and show poor generalization beyond the training scenarios.

DH-PTAM~\cite{DH-PTAM} proposes a stereo event-image parallel tracking and mapping system that replaces hand-crafted features with learned features.
E-RAFT~\cite{e-RAFT} develops a dense optical flow method for event data association, building on the recurrent network architecture of RAFT~\cite{RAFT}, using a volumetric voxel grid representation of event data.
DEVO~\cite{DEVO} extends the DPVO~\cite{DPVO} to accommodate the event modality also through the voxel-based representation like E-RAFT~\cite{e-RAFT}, demonstrating great generalization from synthetic data to seven real-world event-based benchmarks.
RAMP-VO~\cite{rampvo} introduces an end-to-end VO that also builds upon DPVO~\cite{DPVO} using feature encoders to fuse events and image data.
It achieves robust zero-shot transfer performance to real data despite being trained only on synthetic datasets.
Both DEVO~\cite{DEVO} and RAMP-VO~\cite{rampvo} demonstrate the powerful capability of the patch graph and recurrent network architecture from DPVO~\cite{DPVO} and RAFT~\cite{RAFT} for event-based data association.
However, the absolute scale is not observable in these monocular event-only systems without additional information or an IMU. 
They have inherent limitations, with failures manifesting in various ways, such as in low-texture environments, optimization algorithm divergence, and scale ambiguity. 
Visual-IMU integration is the most common solution to address these limitations, which provides scale awareness and continuous estimation with minimal setup.
Nevertheless, integrating a learning-based event network with IMU remains an unexplored territory due to the challenge of efficiently fusing the learning-based event association with IMU. 
This work aims to bridge this gap by proposing a combined learning-optimization framework. 
By leveraging deep learning, it achieves robust event-based data association while simultaneously capitalizing on the complementary strengths of IMU 
The key advantage lies in its ability to train the network independently of IMU data, which not only maintains generalization performance but also improves adaptability across diverse scenarios.

%%%%%%%%%%%%%%%%%%***********************************************************************************************************************************************************%%%%%%%%%%%%%%%%%%
\begin{figure*}[htb]  %%(h 此处（here） t 页顶（top）b 页底（bottom） p 独立一页（page）)
        \centering
        \captionsetup{justification=justified}%图题对齐
        \includegraphics[width=2.0\columnwidth]{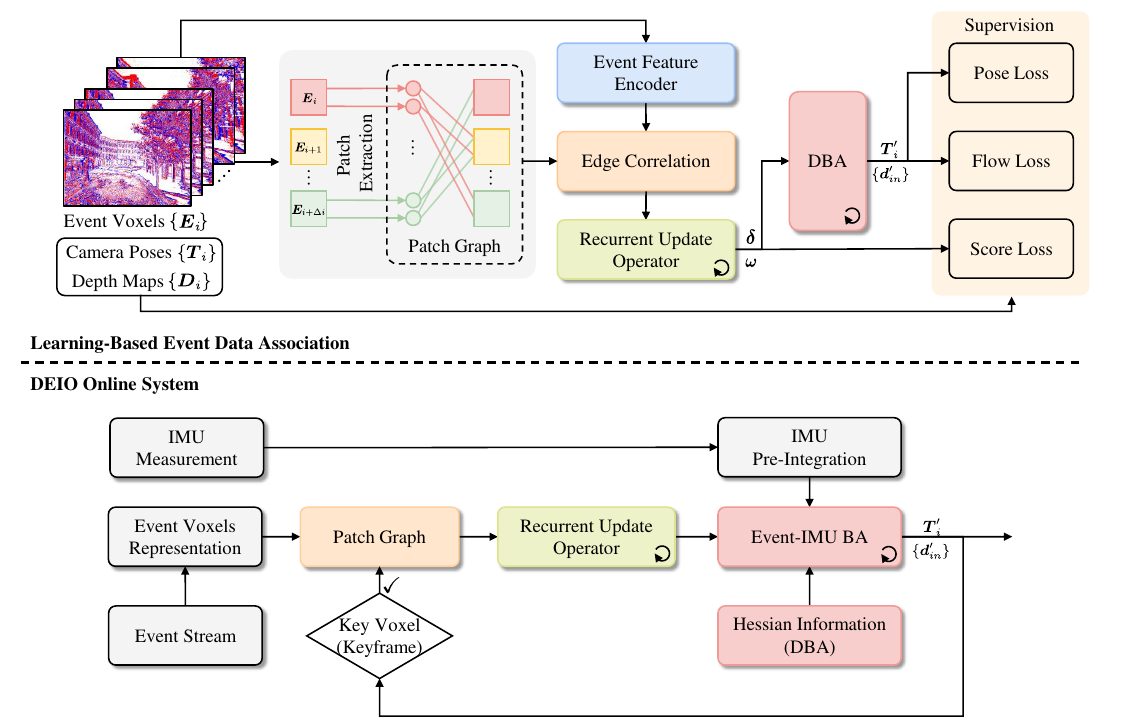}

        \captionsetup{justification=justified}%图题对齐
        \caption{
        Overview of the DEIO system.
        It decouples network training from IMU integration and operates in two phases: offline training and online optimization.
        The main innovations of this work reside in the effective integration of IMU measurements with learning-based methods. 
        During training, a unified event-based optical flow network is trained to provide robust data associations of sparse event patches. 
        At runtime, the Hessian information, derived from the DBA layer in the update operator, is utilized to tightly integrate event patch correspondence with IMU pre-integration through an event patch-based co-visibility factor graph optimization.
        }  %设置图片的一个编号以及为图片添加标题
        \label{figure:framework_overview}
        % \vspace{-1.6em}%调整表格与正文的距离
\end{figure*}%
%%%%%%%%%%%%%%%%%%***********************************************************************************************************************************************************%%%%%%%%%%%%%%%%%%

%%%%%%%%%%%%%%%%%%%%%%%%%%%%%%%%%%%%%%%%%%%%%%%%%%%%%%%%%%%%%%%%%%%%%%%%%%%%%%%%%%%%%%%%%%%%%%%%%%%%%%%%%%%%
\section{Methodology}
\label{section: Methodology}

The overall design of the framework aims to tightly fuse the trainable event-based data association with traditional IMU pre-integration.
\figurename~\ref{figure:framework_overview} depicts the overview of our system.
The main innovations of this work reside in how learning-based methods and traditional graph optimization are fused together.
For the front end, a deep neural network (Section~\ref{subsec: Learning-based Event Data Association}) is utilized to estimate the sparse patch-based correspondence for the optical flow of events.
On the back end, Hessian information derived from the learned DBA layer is tightly integrated with the IMU pre-integration (Section~\ref{subsec: Integrating DBA for Factor Graph}). 
Event reprojections are used to formulate an event patch-based co-visibility factor graph (Section~\ref{subsec: Integrating DBA for Factor Graph}), enabling up-to-scale and robust pose tracking of the entire online system~(Section~\ref{subsec: online eio system}).
This design leverages the representational power of deep neural networks to achieve robust event-based data association while simultaneously harnessing inertial measurement benefits without requiring IMU training data, thereby preserving the generalization capabilities of our DEIO.

%%%%%%%%%%%%%%%%%%%%%%%%%%%%%%%%%%%%%%%%%%%%%%%%%%%%%%%%%%%%%%%%%%%%%%%%%%%%%%%%%%%%%%%%%%%%%%%%%%%%%%%%%%%%
\subsection{Learning-based Event Data Association}
\label{subsec: Learning-based Event Data Association}

%%%%%%%%%%%%%%%%%%%%%%%%%%%%%%%%%%%%%%%%%%%%%%%%%%%%%%%%%%%%%%%%%%%%%%%%%%%%%%%%%%%%%%%%%%%%%%%%%%%%%%%%%%%%
\subsubsection*{\textbf{Event Encoding}}
Event cameras asynchronously capture illumination changes at each pixel location and produce a stream of events. 
Each event is represented as a tuple $(t, x, y, p)$, where $t$ denotes the trigger timestamp in microseconds at the pixel $(x, y)$ and $p_i$ indicates polarity with $p = 1$ for an increase in brightness and $p = -1$ for a decrease. 
The event streams are divided into segments based on a predefined temporal interval $\Delta t$. 
We preprocess the event segment within an interval $[t_i-\Delta t, t_i)$ into a tensor $\boldsymbol{E}_{i}\in \mathbb{R}^{D\times H \times W}$ using the voxel representation~\cite{zhu2019unsupervised}, where $D$ represents the number of discretization steps in time.
Therefore, the event-based optical flow estimation from segment $i$ to segment $j$ fundamentally involves establishing data correspondences between $\boldsymbol{E}_{i}$ and $\boldsymbol{E}_{j}$. 

%%%%%%%%%%%%%%%%%%%%%%%%%%%%%%%%%%%%%%%%%%%%%%%%%%%%%%%%%%%%%%%%%%%%%%%%%%%%%%%%%%%%%%%%%%%%%%%%%%%%%%%%%%%%
\subsubsection*{\textbf{Patch Structure}}
For efficiency, instead of constructing dense correspondences like~\cite{RAFT,e-RAFT}, a patch-based architecture~\cite{DPVO} is adopted to compute the flow for a set of sparse event patches. 
A $p\times p$ event patch, sampled from the event voxel $\boldsymbol{E}$, is represented as a set of pixel coordinates $\boldsymbol{P}=[\boldsymbol{x}, \boldsymbol{y}]\in\mathbb{R}^{p^2 \times 2}$, where all pixels within the patch are assumed to have a constant inverse depth $\boldsymbol{d}\in\mathbb{R}_{+}$. 
The dynamic event patch graph $\mathcal{G}$ is a bipartite graph where each edge is denoted as $[(i,n),j]$, indicating the relationship between event patch $n$ from segment $i$ and the target segment $j$. 
The trajectory of the event patch can be obtained by reprojecting it onto all the connected event segments in the patch graph, thereby forming the sparse event data association.

%%%%%%%%%%%%%%%%%%%%%%%%%%%%%%%%%%%%%%%%%%%%%%%%%%%%%%%%%%%%%%%%%%%%%%%%%%%%%%%%%%%%%%%%%%%%%%%%%%%%%%%%%%%%
\subsubsection*{\textbf{Network Structure}}
The transformation of asynchronous events into a frame-like structure allows compatibility with CNN to model the data association between patches and segments.
The unified network for event-based data association inherits the recurrent network architecture from~\cite{DPVO, DEVO} and consists of three primary components: 
(i) a feature encoder that extracts patch-based event feature representations, including matching and contextual features; 
(ii) a correlation layer that computes the visual similarity between patches and segments, where patches are selected by a patch selector; 
(iii) a recurrent update operator that handles event-patch correspondences coupled with differentiable bundle adjustment (DBA), which estimates the 2D optical flow vector $\boldsymbol{\delta}$ and the confidence weights $\boldsymbol{\omega}$ for each patch onto the target segment in the patch graph $\mathcal{G}$. 

%%%%%%%%%%%%%%%%%%%%%%%%%%%%%%%%%%%%%%%%%%%%%%%%%%%%%%%%%%%%%%%%%%%%%%%%%%%%%%%%%%%%%%%%%%%%%%%%%%%%%%%%%%%%
\subsubsection*{\textbf{Differentiable Bundle Adjustment Layer}}

The DBA layer is employed to bridge the reprojection error and the predicted optical flow of event patches. 
This layer jointly refines camera poses and patch depth across the entire patch graph $\mathcal{G}$ to match the predicted patch optical flow $\boldsymbol{\delta}$ by the following optimization objective:
\begin{equation}
\begin{aligned}
&\left\{\left\{\boldsymbol{T}'_{ji}\},\ \{\boldsymbol{d}'_{in}\right\}\right\}=\arg\min_{\boldsymbol{T}, \boldsymbol{d}}
\sum_{\mathcal{G}} 
\left\Vert\boldsymbol{F}(\boldsymbol{T}_{ji},\boldsymbol{d}_{in})\right\Vert^{2}_{\boldsymbol{\omega}_{inj}}, 
\\\boldsymbol{F}(&\boldsymbol{T}_{ji},\boldsymbol{d}_{in}) = \boldsymbol{\pi}\left(\boldsymbol{T}_{ji} \cdot \boldsymbol{\pi}^{-1}\left(\hat{\boldsymbol{P}}_{in},\boldsymbol{d}_{in}\right)\right) - \left(\hat{\boldsymbol{P}}_{in}+\boldsymbol{\delta}_{inj}\right)
\end{aligned}
\label{equation:e_dba}
\end{equation}
where
$(\boldsymbol{\delta}_{inj},\boldsymbol{\omega}_{inj})$ is the patch-based optical flow field predicted by the event-based recurrent network.
The term $\boldsymbol{\delta}_{inj} \in \mathbb{R}^{2}$ denotes a 2D flow vector that indicates how the reprojection of the event-patch center should be updated, and $\boldsymbol{\omega}_{inj}$ serves as the patch-wise confidence weight of the optical flow.
$\boldsymbol{F}$ is the shorthand to denote the residual term on the patch center coordinates, and $\Vert\cdot\Vert$ is the Mahalanobis distance. 
$\boldsymbol{\pi}$ and $\boldsymbol{\pi}^{-1}$ are the projection and back-projection functions of the event camera. $\boldsymbol{T}_{ji}=\boldsymbol{T}_{j}^{-1}\boldsymbol{T}_{i}$ represents the transformation from frame $i$ to frame $j$, where $\boldsymbol{T}_{i},\boldsymbol{T}_{j}\in\mathbb{SE} (3)$ denote the camera poses in the camera-to-world format. The DBA layer can efficiently backpropagate gradients through Gauss-Newton iterations, thus enabling the pose supervision during network training.

%%%%%%%%%%%%%%%%%%%%%%%%%%%%%%%%%%%%%%%%%%%%%%%%%%%%%%%%%%%%%%%%%%%%%%%%%%%%%%%%%%%%%%%%%%%%%%%%%%%%%%%%%%%%
\subsection{Learnable Hessian Information Extraction}
\label{subsec: Integrating DBA for Factor Graph}

The key component of our proposed learning-optimization combined framework is extracting the information from the learning-based event data association and integrating it with the IMU.
To achieve this, we first linearize the reprojection errors from Eq.~\eqref{equation:e_dba} as follows:
\begin{equation}
\begin{aligned}
% \delta \boldsymbol{F}(\boldsymbol{T}_{ji},\boldsymbol{d}_{i})
\boldsymbol{F}(\boldsymbol{\xi_{ji}}\oplus\boldsymbol{T}_{ji},\boldsymbol{d}_{in}+\Delta\boldsymbol{d}_{in})
&-
\boldsymbol{F}(\boldsymbol{T}_{ji},\boldsymbol{d}_{in})
\\&=
\left[
    \begin{matrix}
    % J_{i} & J_{j} & J_{d_{i}}
    \boldsymbol{J}_{ji} & \boldsymbol{J}_{\boldsymbol{d}_{in}}
    \end{matrix}
\right]
\left[
    \begin{matrix}
    % \boldsymbol{\xi_{i}} & \boldsymbol{\xi_{j}} & \partial d_{i}
    \boldsymbol{\xi}_{ji} \\
    \Delta \boldsymbol{d}_{in}
    \end{matrix}
\right]
\end{aligned}
\label{equation:reprojection_error}
\end{equation}
where $\boldsymbol{\xi}_{ji}$ is the Lie algebras of the updated pose in $\mathbbm{se}(3)$.
$\Delta \boldsymbol{d}_{in}$ denotes the updated state of the inverse depth.
The Jacobians $\boldsymbol{J}_{ji}, \boldsymbol{J}_{\boldsymbol{d}_{in}}$ are the partial derivatives of $\boldsymbol{F}$ with respect to the pose $\boldsymbol{T}_{ji}$ and the inverse depth $\boldsymbol{d}_{in}$, respectively.
An event-patch $\boldsymbol{P}_{in}$ can be reprojected from segment $i$ into segment $j$ follows the warping function:
\begin{equation}
    \boldsymbol{P}^{'}_{jn}=
\left[
    \begin{matrix}
        % x_{jn} &  y_{jn} & z_{jn} & w_{j}
        x_{jn} &  y_{jn} & z_{jn} & 1
    \end{matrix}
\right] ^{T}
=\boldsymbol{T}_{ji} \cdot \boldsymbol{\pi}^{-1}(\hat{\boldsymbol{P}}_{in}, \boldsymbol{d}_{in})
\end{equation}
where $(x_{jn}, y_{jn}, z_{jn})$ is the center of the event patch at segment $j$ in the event camera coordinate.
% The Jacobians of this reprojection can be expressed as:
$\boldsymbol{J}_{ji}$ is expressed as:
\begin{equation}
\resizebox{\columnwidth}{!}{$ %调整大小
\boldsymbol{J}_{ji}=
\left[
\begin{matrix}
\frac{f_{x} }{z_{jn} \cdot \boldsymbol{d}_{in}}  & 0 & -\frac{f_{x} x_{jn} }{z_{jn}^{2} \cdot \boldsymbol{d}_{in}} & -\frac{f_{x} x_{jn} y_{jn}}{z_{jn}^{2}} & f_{x} +\frac{f_{x} x_{jn}^{2}}{z_{jn}^{2}} & -\frac{f_{x} y_{jn}}{z_{jn}} \\
0 & \frac{f_{y} }{z_{jn} \cdot \boldsymbol{d}_{in}}  & -\frac{f_{y} y_{jn} }{z_{jn}^{2} \cdot \boldsymbol{d}_{in} } & -f_{y} -\frac{f_{y} y_{jn}^{2}}{z_{jn}^{2}}  & \frac{f_{y} x_{jn} y_{jn}}{z_{jn}^{2}} & \frac{f_{y} x_{jn}}{z_{jn}} 
\end{matrix}
\right]
$}
\end{equation}
where $f_{x}$ and $f_{y}$ are the given event camera intrinsic parameters.
$\boldsymbol{J}_{\boldsymbol{d}_{in}}$ is given as:
\begin{equation}
    \boldsymbol{J}_{\boldsymbol{d}_{in}}=
\left[
\begin{matrix}
    f_{x}(\frac{\boldsymbol{t}_{ji}[0]}{z_{jn} \cdot \boldsymbol{d}_{in}} - \frac{\boldsymbol{t}_{ji}[2] x_{jn}}{z_{jn}^{2} \cdot \boldsymbol{d}_{in}} )\\ 
    f_{y}(\frac{\boldsymbol{t}_{ji}[1]}{z_{jn} \cdot \boldsymbol{d}_{in} } - \frac{\boldsymbol{t}_{ji}[2] y_{jn}}{z_{jn}^{2} \cdot \boldsymbol{d}_{in} } )
\end{matrix}
\right] 
\end{equation}
where $\boldsymbol{t}_{ji}$ is the translation vector of the related transform between $\boldsymbol{T}_{i}$ and $\boldsymbol{T}_{j}$. 
Therefore, the Hessian matrix of Eq.~\eqref{equation:reprojection_error} can be computed as follows:
\begin{equation}
 \boldsymbol{H}_{ji}=
     \left[
         \begin{matrix}
        \boldsymbol{J}_{ji} & \boldsymbol{J}_{\boldsymbol{d}_{in}}
        \end{matrix}
    \right]^{T}
    \boldsymbol{W}_{inj}
    \left[
         \begin{matrix}
        \boldsymbol{J}_{ji} & \boldsymbol{J}_{\boldsymbol{d}_{in}}
        \end{matrix}
    \right]
\end{equation}
where $\boldsymbol{W}_{inj}=\text{diag}(\boldsymbol{\omega}_{inj})$.
To improve readability, the notation for $[(i,n),j]$ edges in $\boldsymbol{H}_{ji}$ and $\boldsymbol{W}_{inj}$ will be omitted in subsequent equations unless otherwise specified.

% Separating the pose and depth variables, the system can be solved efficiently through Schur complement:
By decoupling the pose and depth variables, the system can be solved efficiently using the Schur complement:
\begin{equation}
\begin{aligned}
& \boldsymbol{H} 
\left[
    \begin{matrix}
    \boldsymbol{\xi}_{ji} \\
    \Delta \boldsymbol{d}_{in}
    \end{matrix}
\right] =
-\left[
     \begin{matrix}
    \boldsymbol{J}_{ji} & \boldsymbol{J}_{\boldsymbol{d}_{in}}
    \end{matrix}
\right]^{T} \boldsymbol{W} \boldsymbol{F}(\boldsymbol{T}_{ji},\boldsymbol{d}_{in}) \\
&
\left[
    \begin{matrix}
    \boldsymbol{B}     & \boldsymbol{E} \\
    \boldsymbol{E}^{T} & \boldsymbol{C}
    \end{matrix}
\right]
\left[
    \begin{matrix}
    \boldsymbol{\xi}_{ji} \\
    \Delta \boldsymbol{d}_{in}
    \end{matrix}
\right]
=
\left[
    \begin{matrix}
    \boldsymbol{v} \\
    \boldsymbol{u}
    \end{matrix}
\right] 
\end{aligned}
\end{equation}
Therefore, the following equation can be obtained:
\begin{equation}
    \boldsymbol{\xi_{ij}}=\underbrace{[\boldsymbol{B}-\boldsymbol{E} \boldsymbol{C}^{-1}\boldsymbol{E}^{T}]^{-1}}_{\boldsymbol{H}_{g}} \underbrace{(\boldsymbol{v}-\boldsymbol{E}\boldsymbol{C}^{-1}\boldsymbol{u})}_{\boldsymbol{V}_{g}}
\label{equation:update_pose}
\end{equation}
where
$\boldsymbol{v}$ is a $6K \times 1$ vector (with 6 DoF pose and totally $K=10$ keyframe), and $\boldsymbol{u}$ is an $NK \times 1$ vector (with totally number of $N=96$ event patches for each event segment).
$\boldsymbol{B}$ is the matrix with size of $60 \times 60$, $\boldsymbol{E}$ is the residual matrix with size of $60 \times 960$, and $\boldsymbol{C}$ is a diagonal matrix with size of $960 \times 960$.
A damping factor of $10^{-4}$ is also applied to $\boldsymbol{C}$ as~\cite{DPVO}.
These matrices establish an interframe pose constraint (represented by $\boldsymbol{H}_{g}$ and $\boldsymbol{V}_{g}$) that integrates the DBA information.
% After updating the camera poses $\boldsymbol{T}_{j}^{'}=\boldsymbol{T}^{'}_{ji} \cdot \boldsymbol{T}_{i}$, $\boldsymbol{T}_{ji}=\text{Exp}(\boldsymbol{\xi}_{ji})$, the inverse depth of each event patch can be updated as:
After updating the camera poses $\boldsymbol{T}'_{ji}=\text{Exp}(\boldsymbol{\xi}_{ji})\boldsymbol{T}_{ji}$, the inverse depth of each event patch can be updated as:
\begin{equation}
\begin{aligned}
&    \boldsymbol{d}'_{in}=\Delta \boldsymbol{d}_{in}+\boldsymbol{d}_{in}  \\
&     \Delta \boldsymbol{d}_{in}=\boldsymbol{C}^{-1}(\boldsymbol{u}-\boldsymbol{E}^{T}\boldsymbol{\xi}_{ji})
\label{equation:update_depth}
\end{aligned}
\end{equation}

The calculations of Eq.~\eqref{equation:update_pose} and Eq.~\eqref{equation:update_depth} can be efficiently performed in parallel on GPU with CUDA acceleration.
All the Hessian information $\boldsymbol{H}_{g}$ and the corresponding $\boldsymbol{V}_{g}$, derived from the co-visibility graph, are integrated into the factor graph where they are optimized on the CPU.
The DBA contributes extensive geometric information, incorporating learned uncertainties, to the factor graph.
The optimization results (updated poses and depths) are then iteratively fed back to refine the event-based optical flow network. 
This recurrent process enhances the robustness and accuracy of the overall system.

%%%%%%%%%%%%%%%%%%%%%%%%%%%%%%%%%%%%%%%%%%%%%%%%%%%%%%%%%%%%%%%%%%%%%%%%%%%%%%%%%%%%%%%%%%%%%%%%%%%%%%%%%%%%
\subsection{Graph-Based Event-IMU Combined Bundle Adjustment}
\label{subsec: Graph-based DBA and IMU Optimization}

Unlike end-to-end approaches that use deep networks to fuse the features from two modalities (visual and IMU) and predict poses directly, our DEIO combines the neural networks with event-inertial bundle adjustment.
To this end, we design a learning-optimization combined framework that tightly integrates the Hessian information from DBA and IMU pre-integration within keyframe-based sliding window optimization.
The full state vector of the $k$th keyframe in the sliding window (with the total number of keyframes $K=10$ in our implementation), is defined as: 
\begin{equation}
\boldsymbol{\chi}=[\boldsymbol{T}^{w}_{b_{k}}, \boldsymbol{v}^{w}_{b_{k}}, \boldsymbol{b}_{a_{k}}, \boldsymbol{b}_{g_{k}}], k=1,2,3,...
\label{state}
\end{equation}
where
$\boldsymbol{T}^{w}_{b_{k}}=
\left[
    \begin{matrix}
        \boldsymbol{R}^{w}_{b_{k}} & \boldsymbol{t}^{w}_{b_{k}}\\
        0 & 1
    \end{matrix}
\right] \in \mathbb{SE} (3)$ 
is the pose of the body (IMU) frame in the world frame, given by the translation $\boldsymbol{t}^{w}_{b_{k}}$ and rotation matrix $\boldsymbol{R}^{w}_{b_{k}}$.
$\boldsymbol{v}^{w}_{b_{k}}$ is the velocity of the IMU in the world frame.
$\boldsymbol{b}_{a_{k}}$ and $\boldsymbol{b}_{g_{k}}$ are the accelerometer bias and  gyroscope bias, respectively.

We solve the state estimation problem by constructing a factor graph with the GTSAM library and optimizing it with the Levenberg-Marquardt.
The cost function, which minimizes residuals from various factors corresponding to different aspects of data constraints, can be written as:
\begin{equation}
\begin{aligned}
       \boldsymbol{J}(\boldsymbol{\chi})=
       ||\boldsymbol{r}^{k}_{\text{event}}||^{2}_{W^{k}_{\text{event}}}
       +\sum_{k=0}^{K-1}||\boldsymbol{r}^{k}_{\text{imu}}||^{2}_{W^{k}_{\text{imu}}}
       +||\boldsymbol{r}_{m}||^{2}_{W_{m}}
\label{Joint_nonlinear_optimization}
\end{aligned}
\end{equation}
Eq.~\eqref{Joint_nonlinear_optimization} contains 
the event-based residuals $\boldsymbol{r}^{k}_{\text{event}}$ with weight $W^{k}_{\text{event}}$,
the IMU pre-integration residuals $\boldsymbol{r}^{k}_{\text{imu}}$ with weight $W^{k}_{\text{imu}}$, and
the marginalization residuals $\boldsymbol{r}_{m}$ with weight $W_{m}$,

Given the Hessian information $\boldsymbol{H}_{g}$ and the corresponding $\boldsymbol{V}_{g}$, the event residual factor can be written as:
\begin{equation}
\begin{aligned}
    \boldsymbol{r}^{k}_{\text{event}}=&\frac{1}{2}
\left[
\begin{matrix}
 \boldsymbol{\xi}^{w}_{e_{0}} & \boldsymbol{\xi}^{w}_{e_{1}}& \cdots& \boldsymbol{\xi}^{w}_{e_{k}}    
\end{matrix}
\right] 
    \boldsymbol{H}_{g} 
\left[
\begin{matrix}
 \boldsymbol{\xi}^{w}_{e_{0}} \\ \boldsymbol{\xi}^{w}_{e_{1}} \\ \vdots \\ \boldsymbol{\xi}^{w}_{e_{k}}    
\end{matrix}
\right] \\
    &-
\left[
    \begin{matrix}
     \boldsymbol{\xi}^{w}_{e_{0}} & \boldsymbol{\xi}^{w}_{e_{1}}& \cdots& \boldsymbol{\xi}^{w}_{e_{k}}    
    \end{matrix}
\right] 
    \boldsymbol{V}_{g}
\end{aligned}
\end{equation}
where $\boldsymbol{\xi}^{w}_{e_{k}}=\boldsymbol{\xi}^{e}_{b} \cdot \boldsymbol{\xi}^{w}_{b_{k}}$, and $ \boldsymbol{\xi}^{w}_{b_{k}} =\log_{\mathbb{SE} (3)}(\boldsymbol{T}^{w}_{b_{k}}) $.
$\boldsymbol{\xi}^{w}_{e_{k}}$ and $\boldsymbol{\xi}^{w}_{b_{k}}$ are the Lie algebras of the event camera pose and IMU pose in $k^{th}$ keyframe, respectively.
$\boldsymbol{\xi}^{e}_{b}$ is the extrinsics between the event camera and IMU.

Eventually, the IMU residual factor can be derived as follows:
\begin{equation}
\begin{split}
\label{IMU_residual}
\boldsymbol{r}^{k}_{\text{imu}} =
\left[
\begin{array}{c}
\boldsymbol{R}^{b_{k}}_{w}(\boldsymbol{t}^{w}_{b_{k+1}}-\boldsymbol{t}^{w}_{b_{k}}-\boldsymbol{v}^{w}_{b_{k}}\Delta t - \frac{1}{2}\boldsymbol{g}^{w}\Delta t^{2})- \hat{\boldsymbol{\alpha}}^{b_{k}}_{b_{k+1}}  \\
\boldsymbol{R}^{b_{k}}_{w}(\boldsymbol{v}^{w}_{b_{k+1}}-\boldsymbol{v}^{w}_{b_{k}}-\boldsymbol{g}^{w}\Delta t)- \hat{\boldsymbol{\beta}}^{b_{k}}_{b_{k+1}}\\
2\left[ (\boldsymbol{q}^{w}_{b_{k}})^{-1} \otimes \boldsymbol{q}^{w}_{b_{k+1}} \otimes (\hat{\boldsymbol{\gamma}}^{b_{k}}_{b_{k+1}})^{-1} \right]_{xyz} \\
\boldsymbol{b}_{a_{k+1}}-\boldsymbol{b}_{a_{k}}\\
\boldsymbol{b}_{g_{k+1}}-\boldsymbol{b}_{g_{k}}\\
\end{array}
\right]
\end{split}
\end{equation}
where $\boldsymbol{\alpha}^{b_{k}}_{b_{k+1}}$, $\boldsymbol{\beta}^{b_{k}}_{b_{k+1}}$, $\boldsymbol{\gamma}^{b_{k}}_{b_{k+1}}$ are the IMU pre-integration term~\cite{GWPHKU:PL-EVIO};
$\boldsymbol{g}^{w}$ is the gravity vector;
$\Delta t$ is the time interval between keyframe $k$ and $k+1$;
$\boldsymbol{q}^{b_{k}}_{w}$ is the quaternion of the corresponding rotation matrix $\boldsymbol{R}^{b_{k}}_{w}$ with $[\cdot]_{xyz}$ extracts the vector portion.

%%%%%%%%%%%%%%%%%%%%%%%%%%%%%%%%%%%%%%%%%%%%%%%%%%%%%%%%%%%%%%%%%%%%%%%%%%%%%%%%%%%%%%%%%%%%%%%%%%%%%%%%%%%%
\subsection{Online EIO System}
\label{subsec: online eio system}
\begin{figure*}[htb]  %%(h 此处（here） t 页顶（top）b 页底（bottom） p 独立一页（page）)
        \centering        
        \includegraphics[width=2.0\columnwidth]{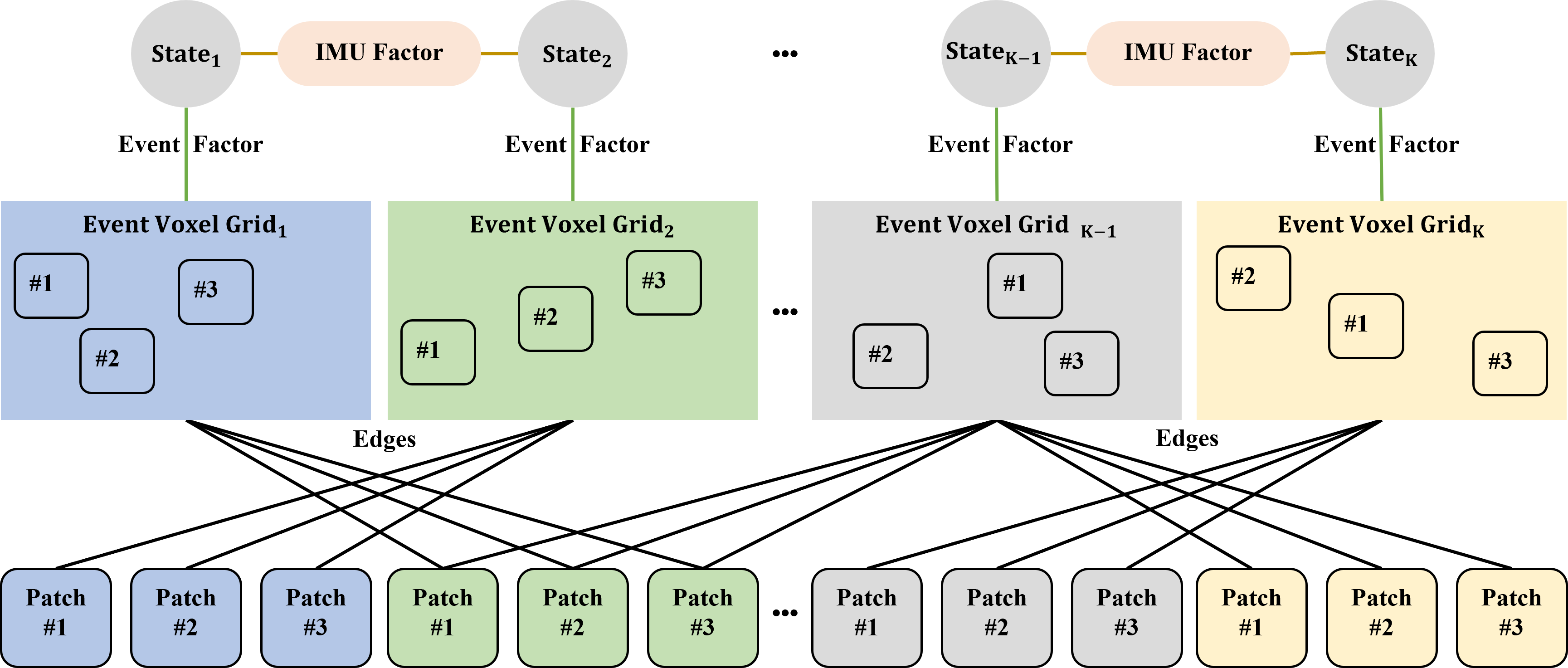}
        % \captionsetup{justification=justified}%图题对齐
        \caption{
        Patch-based co-visibility factor graph for event-IMU combined bundle adjustment.
        }  %设置图片的一个编号以及为图片添加标题
        \label{figure:patch_graph}
        % \vspace{-1.0em}%调整表格与正文的距离
\end{figure*}

As illustrated in Fig.~\ref{figure:patch_graph}, our DEIO maintains a patch-based co-visibility factor graph that takes the raw event stream and IMU data as input and performs 6 DoF pose estimation.

%%%%%%%%%%%%%%%%%%***********************************************************************************************************************************************************%%%%%%%%%%%%%%%%%%
%%%%%%%%%%%%%%%%%%***********************************************************************************************************************************************************%%%%%%%%%%%%%%%%%%

\subsubsection*{\textbf{Initialization}}
We use 8 event segments for event-only initialization.
The new event patches are added until 8 event segments are accumulated, and then run 12 iterations of the update operator, followed by two event-only bundle adjustment iterations.
Since the IMU provides scale awareness and the gravity direction, it is essential to initialize the states in a metric-scale local frame for better convergence.
Similar to the initialization procedure from~\cite{GWPHKU:VINS-MONO,GWPHKU:VINS-MONO-initialization}, where the vision-only structure from motion (SfM) is replaced by the event-only bundle adjustment.
After the event-only initialization, we employ the IMU pre-integration to establish the up-to-scale camera pose.
Following linear alignment and gravity refinement, the scale between the IMU and the event camera is restored, with further correction applied to the depth of the event-based patch.
Subsequently, the event-IMU bundle adjustment (Section~\ref{subsec: Graph-based DBA and IMU Optimization}) is iteratively performed by updating the co-visibility graph based on the event-inertial initialization results.
We also found that with the accurate pose estimation provided by event-based SfM, event-inertial initialization can be easily accomplished along with high-precision gravity recovery.
For event-only initialization, sufficient camera motion is essential. 
To ensure this, we only accumulate event segments that have an average optical flow magnitude of at least 8 pixels. 
This magnitude is estimated after a single update iteration from the preceding segment.
In addition, sufficient excitation is also required from the IMU, with the gyroscope’s variance exceeding 0.25.

\subsubsection*{\textbf{Co-visibility Patch Graph Management}}
When a new event patch is added, the edges between that event patch and the previous $r$ keyframes ($r=13$ in our implementations) are added to the patch graph.
After appending these new edges (with the predicted event-based optical flow field $(\boldsymbol{\delta}_{inj},\boldsymbol{\omega}_{inj})$), the update operation of the patch graph is performed, including the recurrent update operator and the 2 event-IMU bundle adjustment iterations.
After that, the keyframe selection is performed to maintain the keyframe-based sliding window.

\subsubsection*{\textbf{Keyframe Strategy}}
The most recent 3 event segments are always considered keyframes.
After each update operation of the co-visible graph, the optical flow magnitude between the $(t-5)$th and $(t-3)$th keyframe will be calculated.
If this magnitude is less than 60 pixels, the $(t-4)$th keyframe, along with all associated edges and patch-based features, will be removed.
Additionally, the event patches and keyframes are discarded once they fall outside the optimization window.
The oldest keyframe that is abandoned will be marginalized, along with its associated IMU pre-integration, to construct the marginalization factor $\boldsymbol{r}_{m}$ in Eq.~\eqref{Joint_nonlinear_optimization}.
% And we roll up window states to conserve memory.

%%%%%%%%%%%%%%%%%%%%%%%%%%%%%%%%%%%%%%%%%%%%%%%%%%%%%%%%%%%%%%%%%%%%%%%%%%%%%%%%%%%%%%%%%%%%%%%%%%%%%%%%%%%%
\section{Experiments}
\label{section:Experiments}

We conduct quantitative and qualitative evaluations of our DEIO across \textit{ten} challenging real-world datasets with varying camera resolution and diversity scenarios on different platforms.
Specifically, in Section~\ref{section:Comparisons with SOTA Methods in Challenge Benchmarks}, we compare DEIO with baseline methods across multiple challenging event datasets, showcasing its superior performance and exceptional generalization capabilities. 
While in Section~\ref{section:driving scenarios} and~\ref{section:drone scenarios}, we assess the performance of our DEIO in night driving scenarios and indoor low-light quadrotor flights, respectively. 
DEIO is compared against over 20 baseline methods from various literature, including both event-based and image-based approaches. 
Finally, Section~\ref{section:Discussion} provides a detailed qualitative analysis and time efficiency evaluation of DEIO.

The event-based data association network is implemented using PyTorch and executed on an NVIDIA RTX-3090 GPU. 
Leveraging the strong generalization capabilities of the recurrent optical flow pipeline~\cite{RAFT, Droid-slam, DPVO, DEVO}, we employ network weights~\cite{DEVO} pre-trained exclusively on the Event-based TartanAir~\cite{Tartanair} dataset, where events are synthesized using the ESIM~\cite{rebecq2018esim} simulator.

%%%%%%%%%%%%%%%%%%***********************************************************************************************************************************************************%%%%%%%%%%%%%%%%%%
\begin{table*}[htb]
        \begin{center}
        \captionsetup{justification=justified}
        \caption{
        Accuracy comparison [MPE(\%)] of our DEIO with other monocular event-based baselines in DAVIS240c dataset~\cite{GWPHKU:event-camera-dataset_davis240c}.
        The estimated trajectory is aligned with the ground truth over the first 5 seconds.
        % Unit: MPE(\%).
        % 0.06 means the average error is 0.06 m for 100 m motion. 
        % Aligning the first 5 seconds of the estimated trajectory with the ground truth. 
        }
        \label{table:davis240c_Comparison}
        \resizebox{2.0\columnwidth}{!}
        { 
        \begin{threeparttable}
        \renewcommand{\arraystretch}{1.2}
        \setlength{\tabcolsep}{1.0mm}
        \begin{tabular}{c|c|cccccccc|c} 
        \hline
        Methods & Modality & boxes\_translation & hdr\_boxes & boxes\_6dof & dynamic\_translation & dynamic\_6dof & poster\_translation & hdr\_poster & poster\_6dof & Average \\
        \hline
        Zhu et al.~\cite{GWPHKU:Event-based-visual-inertial-odometry} & E+I & 2.69 & 1.23 & 3.61 & 1.90 & 4.07 & 0.94 & 2.63 & 3.56 & 2.58 \\
        Henri et al.~\cite{GWPHKU:ETH-EVIO}                             & E+I & 0.57 & 0.92 & 0.69 & 0.47 & 0.54 & 0.89 & 0.59 & 0.82 & 0.69 \\
        Ultimate-SLAM~\cite{GWPHKU:Ultimate-SLAM}      & E+I & 0.76 & 0.67 & 0.44 & 0.59 & 0.38 & 0.15 & 0.49 & 0.30 & 0.47 \\
        Ultimate-SLAM~\cite{GWPHKU:Ultimate-SLAM}          & E+F+I & 0.27 & 0.37 & 0.30 & 0.18 & 0.19 & 0.12 & 0.31 & 0.28 & 0.25 \\
        Jung et al.~\cite{jung2020constrained}          & E+I & 1.50 & 2.45 & 2.88 & 4.92 & 6.23 & 3.43 & 2.38 & 2.53 & 3.92 \\
        Jung et al.~\cite{jung2020constrained}          & E+F+I & 1.24 & 1.15 & 0.98 & 0.89 & 0.98 & 1.83 & 0.57 & 0.97 & 1.07 \\
        HASTE-VIO~\cite{HASTE-VIO}                              & E+I & 2.55 & 1.75 & 2.03 & 1.32 & 0.52 & 1.34 & 0.57 & 1.50 & 1.45 \\
        EKLT-VIO~\cite{EKLT-VIO}                                & E+F+I & 0.48 & 0.46 & 0.84 & 0.40 & 0.79 & 0.35 & 0.65 & 0.35 & 0.54 \\
        Dai et al.~\cite{IROS2022_EVIO}                               & E+I & 1.0 & 1.8 & 1.5 & 0.9 & 1.5 & 1.9 & 2.8 & 1.2 & 1.56 \\
        Mono-EIO~\cite{GWPHKU:MyEVIO}                           & E+I & 0.34 & 0.40 & 0.61 & 0.26 & 0.43 & 0.40 & 0.40 & 0.26 & 0.39 \\
        Kai et al.~\cite{tang2024monocular}                     & E+I & 0.36 & 0.31 & 0.32 & 0.59 & 0.49 & 0.23 & 0.18 & 0.31 & 0.35 \\
        PL-EVIO~\cite{GWPHKU:PL-EVIO}                           & E+F+I & 0.06 & 0.10 & 0.21 & 0.24 & 0.48 & 0.54 & 0.12 & 0.14 & 0.24 \\
        Lee et al.~\cite{lee2023event}                                & E+F+I & 0.74 & 0.69 & 0.77 & 0.71 & 0.86 & 0.28 & 0.52 & 0.59 & 0.65 \\
        EVI-SAM~\cite{GWPHKU:EVI-SAM}                           & E+F+I & 0.11 & 0.13 & 0.16 & 0.30 & 0.27 & 0.34 & 0.15 & 0.24 & 0.21 \\
        DPVO~\cite{DPVO}                                        & F  & \textbf{0.02} & 0.71 &0.59 & 0.09  & 0.05 &0.20  &0.49  &0.44  &0.32\\
        DBA-Fusion~\cite{DBA-Fusion}                            & F+I &0.07 & 0.27 &0.10 & 0.56& 0.11 &0.13  &0.38  & 0.19 &0.23 \\
        DEVO~\cite{DEVO}                                        & E  &0.06 & \textbf{0.06} & 0.71 &0.09 &0.08  &0.06 & 0.14 & 0.44 &0.21\\      
   \textbf{DEIO}                                                 & E+I &0.07 & 0.09 & \textbf{0.05} & \textbf{0.06} & \textbf{0.04} & \textbf{0.04} & \textbf{0.06} & \textbf{0.08} & \textbf{0.06}\\
        \hline       
        \end{tabular}
        \end{threeparttable} 
        }
        \end{center}
        \vspace{-1.5em}%调整表格与正文的距离
\end{table*}
%%%%%%%%%%%%%%%%%%***********************************************************************************************************************************************************%%%%%%%%%%%%%%%%%%

%%%%%%%%%%%%%%%%%%***********************************************************************************************************************************************************%%%%%%%%%%%%%%%%%%
\begin{figure*}[htb]  %%(h 此处（here） t 页顶（top）b 页底（bottom） p 独立一页（page）)
        \centering        
        \captionsetup{justification=justified}%图题对齐
        \includegraphics[width=2.0\columnwidth]{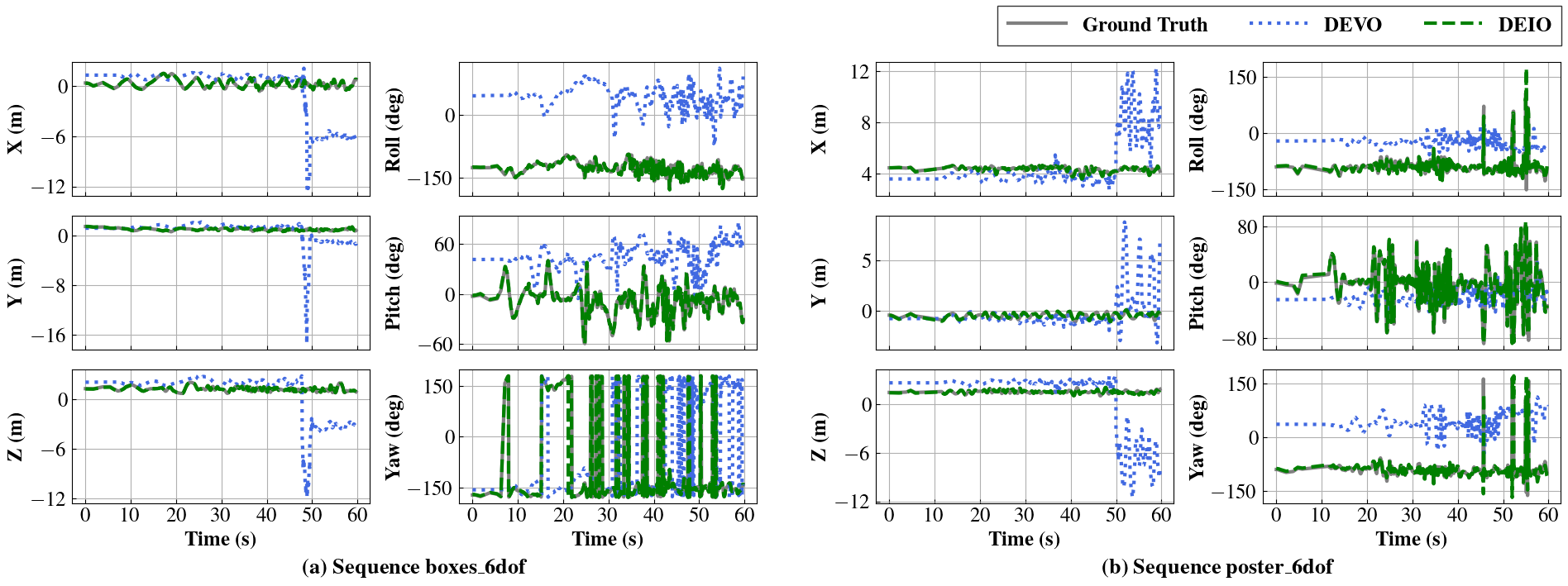}
        \caption{
        Comparison of the estimated position (X, Y, Z) and orientation (Roll, Pitch, Yaw) results of our DEIO with DEVO~\cite{DEVO} in the sequence of (a) boxes\_6dof and (b) poster\_6dof from the DAVIA240c dataset~\cite{GWPHKU:event-camera-dataset_davis240c}. 
        The DEIO efficiently converts scale ambiguity and aligns closely with the ground truth trajectory.
        }  %设置图片的一个编号以及为图片添加标题
        \label{figure:comparison_devo_davis240c}
        % \vspace{-1.5em}%调整表格与正文的距离
\end{figure*}%
%%%%%%%%%%%%%%%%%%***********************************************************************************************************************************************************%%%%%%%%%%%%%%%%%%

%%%%%%%%%%%%%%%%%%%%%%%%%%%%%%%%%%%%%%%%%%%%%%%%%%%%%%%%%%%%%%%%%%%%%%%%%%%%%%%%%%%%%%%%%%%%%%%%%%%%%%%%%%%%
\subsection{Comparisons with SOTA Methods in Challenge Benchmarks}
\label{section:Comparisons with SOTA Methods in Challenge Benchmarks}

To ensure a fair comparison, a consistent trajectory alignment protocol is required.
Therefore, we employ different alignment ways and evaluation criteria according to the compared baseline methods.
For example, regarding the DAVIS240c dataset~\cite{GWPHKU:event-camera-dataset_davis240c}, the estimated and ground-truth trajectories are aligned using a 6-DOF transformation (in $\mathbb{SE}(3)$) over the 5 seconds, and we calculate the mean position error (MPE) as a percentage of the total traveled distance of the ground truth as the evaluation criterion.
In contrast, for the stereo HKU dataset~\cite{ESVIO}, although MPE is also used as the metric, we align the estimated trajectories and ground truth in $\mathbb{SE}(3)$ using all available data, rather than limiting alignment to the 5-second segment, to maintain consistency with the baseline methods.
While for the TUM-VIE dataset~\cite{TUM-VIE}, the baseline methods use the root mean squared error (RMSE) and the Absolute Trajectory Errors (ATE) based on the estimated camera pose as the evaluation criterion.
To avoid confusion, we annotate the trajectory alignment protocol in the footnotes of each table, ensuring that all methods within the same table adhere to the same alignment protocol.
The notations E, F, and I in each table represent the use of event, frame, and IMU, respectively. %, aiding in a clearer performance comparison across different modalities.

%%%%%%%%%%%%%%%%%%%%%%%%%%%%%%%%%%%%%%%%%%%%%%%%%%%%%%%%%%%%%%%%%%%%%%%%%%%%%%%%%%%%%%%%%%%%%%%%%%%%%%%%%%%%
\subsubsection*{\textbf{DAVIS240C}~\cite{GWPHKU:event-camera-dataset_davis240c}} 

This is the most classical monocular event-based SLAM benchmark, which provides event and image data with a resolution of 240$\times$180 pixels, as well as IMU data and 200 Hz ground truth poses. 
It contains extremely fast 6-Dof motion and scenes with HDR. 
The baseline results (except for DPVO, DBA-Fusion, and DEVO) are directly taken from the original works, which also employed the same trajectory alignment protocol.
% Additionally, the baselines we validated used the default parameters provided by the original works.
As shown in Table~\ref{table:davis240c_Comparison}, EVI-SAM achieves the best performance among the non-learning methods. 
In contrast, learning-based methods (such as DEVO) can achieve performance comparable to EVI-SAM (which combines both direct and feature-based methods) using purely event sensors, highlighting the effectiveness and strength of learning-based approaches. 
Meanwhile, our learning-optimization combined method exhibits significantly superior performance compared to other learning-based methods (DPVO, DBA-Fusion, and DEVO).
Compared to DEVO, our proposed DEIO reduces the pose tracking error by up to 71\%, owing to the effective integration of learning-based and traditional optimization methods.
Furthermore, Fig.~\ref{figure:comparison_devo_davis240c} illustrates the estimated trajectory of our DEIO, which shows less drift compared to DEVO, verifying the contribution of IMU integration to maintain low-drifting, metric-scale pose estimation.

%%%%%%%%%%%%%%%%%%***********************************************************************************************************************************************************%%%%%%%%%%%%%%%%%%
\begin{table*}[htb] 
        \renewcommand\arraystretch{1.2} %调整高度
        \begin{center}
        \captionsetup{justification=justified}
        \caption{
        Accuracy comparison [MPE(\%)] of our dEIO with other image/event-based baselines in Mono-HKU dataset~\cite{GWPHKU:MyEVIO}.
        The estimated trajectory is aligned with the ground truth over the first 5 seconds.
        % Unit: MPE(\%).
        % Aligning the first 5 seconds of the estimated trajectory with the ground truth.
        }
        \label{table:mono_hku_Comparison_transposed}
        \resizebox{\columnwidth*2}{!}
        { 
        \begin{threeparttable}
        \begin{tabular}{c|c|c|cccccccccc|c} 
        \hline  %插入水平分割线
        Resolution & \makecell{Methods} & \makecell{Modality} & \makecell{vicon\\\_hdr1} & \makecell{vicon\\\_hdr2} & \makecell{vicon\\\_hdr3} & \makecell{vicon\\\_hdr4} & \makecell{vicon\\\_darktolight1} & \makecell{vicon\\\_darktolight2} & \makecell{vicon\\\_lighttodark1} & \makecell{vicon\\\_lighttodark2} & \makecell{vicon\\\_dark1} & \makecell{vicon\\\_dark2} & \makecell{Average} \\

        \hline
        \multirow{12}*{\makecell{DAVIS346 \\ (346$\times$260)}} 
        & ORB-SLAM3~\cite{ORB-SLAM3}        &F &0.32 &0.75 & 0.60 &0.70 &0.75 &0.76 &0.41 &0.58 &\textit{failed} &0.60  &0.61 \\
        & VINS-MONO~\cite{GWPHKU:VINS-MONO} & F+I & 0.96 & 1.60 & 2.28 & 1.40 & 0.51 & 0.98 & 0.55 & 0.55 & 0.88 & 0.52 &1.02  \\
        & DBA-Fusion~\cite{DBA-Fusion}      & F+I & 0.32 & 0.41 & \textit{failed}  & \textit{failed}  & 0.72 & 0.55 &  \textit{failed}  &2.65 & 3.32 &  \textit{failed}   & 1.33 \\
        & \multirow{2}*{Ultimate-SLAM~\cite{GWPHKU:Ultimate-SLAM}}     &E+I  &1.49 & 1.28 & 0.66 & 1.84 &1.33 &1.48 &1.79 &1.32 &1.75 &1.10  & 1.40\\
        &  ~                                & E+F+I       &2.44 &1.11 &0.83 &1.49 &1.00 &0.79 &0.84 &1.49 &3.45 &0.63 & 1.41\\
        & Mono-EIO \cite{GWPHKU:MyEVIO}     &E+I & 0.59 & 0.74 & 0.72 & 0.37 & 0.81 & 0.42 & 0.29 & 0.79 & 1.02 & 0.49 & 0.62  \\
        & PL-EIO~\cite{GWPHKU:PL-EVIO}      & E+I & 0.57 & 0.54 & 0.69 & 0.32 & 0.66 & 0.51 & 0.33 & 0.53 & 0.35 & 0.38 & 0.49 \\
        & PL-EVIO~\cite{GWPHKU:PL-EVIO}     &F+E+I & 0.17 & 0.12 & 0.19 & 0.11 & 0.14 & 0.12 & 0.13 & 0.16 & 0.43 & 0.47 & 0.20 \\
        & DEVO~\cite{DEVO}                  &E &\textbf{0.11} & \textbf{0.07} & \textbf{0.12} & 0.07 &0.97 & 0.12 &0.15 & \textbf{0.12} & 0.07 &\textbf{0.07}  & 0.19 \\
        &  \textbf{DEIO}             & E+I &0.14 &0.09 &0.16 & \textbf{0.07}  &\textbf{0.11} &\textbf{0.10} &\textbf{0.11} & 0.13 & \textbf{0.05} & 0.08  & \textbf{0.10}\\
        \hline        
        \end{tabular}
        \end{threeparttable} 
        }
        \end{center}
         % \vspace{-1.5em}%调整表格与正文的距离
\end{table*}
%%%%%%%%%%%%%%%%%%***********************************************************************************************************************************************************%%%%%%%%%%%%%%%%%%

%%%%%%%%%%%%%%%%%%***********************************************************************************************************************************************************%%%%%%%%%%%%%%%%%%
\begin{figure*}[htb]  %%(h 此处（here） t 页顶（top）b 页底（bottom） p 独立一页（page）)
        \centering        
        \captionsetup{justification=justified}%图题对齐
        \includegraphics[width=2.0\columnwidth]{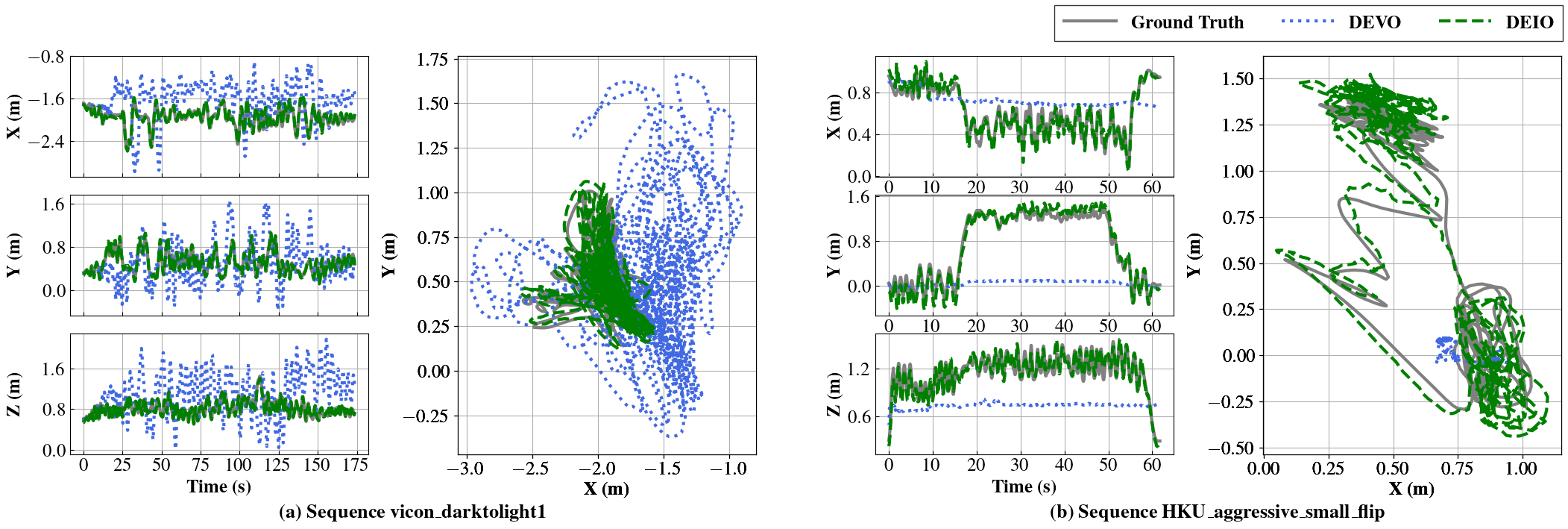}
        \caption{
        Comparison of the estimated trajectories (in X, Y, Z, and XY-plane) with DEVO~\cite{DEVO} in the Mono \& Stereo HKU dataset~\cite{GWPHKU:MyEVIO, ESVIO}. 
        The DEIO seamlessly addresses scale ambiguity and demonstrates precise alignment with the ground truth trajectory.
        In contrast, the baseline estimates exhibit significant scale discrepancies:
        (a) The baseline trajectory suffers from drift and an overestimated scale.
        (b) The baseline trajectory shows an underestimated scale.
        }  %设置图片的一个编号以及为图片添加标题
        \label{figure:comparison_hku}
\end{figure*}%
%%%%%%%%%%%%%%%%%%***********************************************************************************************************************************************************%%%%%%%%%%%%%%%%%%

%%%%%%%%%%%%%%%%%%%%%%%%%%%%%%%%%%%%%%%%%%%%%%%%%%%%%%%%%%%%%%%%%%%%%%%%%%%%%%%%%%%%%%%%%%%%%%%%%%%%%%%%%%%%
\subsubsection*{\textbf{Monocular HKU-dataset}~\cite{GWPHKU:MyEVIO}} 

This dataset is collected using DAVIS346 (346$\times$260, event, image, and IMU sensor), along with a motion capture system (VICON) to obtain pose ground truth.
To mitigate the significant impact of the active infrared (IR) emitters from the VICON cameras on the event camera, an IR filter lens is used to eliminate IR interference. 
As a result, this dataset shows more pronounced thermal noise compared to others.
Furthermore, all sequences are recorded in HDR scenarios under very low illumination or strong illumination changes by switching the strobe flash on and off. 
Table~\ref{table:mono_hku_Comparison_transposed} demonstrates that DEIO outperforms all the event-based methods and decreases the average pose tracking error by at least 47\%.
As illustrated in Fig.~\ref{figure:comparison_hku}(a), the estimated trajectory of DEVO suffers from significant scale loss because the absolute scale cannot be observed in monocular event-only odometry.
This is evident from the projection of the estimated trajectory onto the XY plane, where the blue curve (baseline method) significantly deviates from the ground truth due to the scale loss.
In contrast, our DEIO, despite also being based on a monocular setup, effectively overcomes scale ambiguity and aligns closely with the ground truth trajectory. %, thanks to the compensation provided by the IMU.
This improvement is attributed to the effective compensation provided by the IMU.

%%%%%%%%%%%%%%%%%%***********************************************************************************************************************************************************%%%%%%%%%%%%%%%%%%
\begin{table*}[htb]
    \renewcommand\arraystretch{1.2}
    \begin{center}
    \captionsetup{justification=justified}
    \caption{
    Accuracy comparison [MPE(\%)] of our DEIO with other image/event-based baselines in Stereo-HKU dataset~\cite{ESVIO}.
    The entire sequence of estimated poses is aligned with the ground truth trajectory.
    The baseline results (DPVO, DEVO) are taken from~\cite{DEVO}, and the results of Kai et al. are taken from~\cite{tang2024monocular}.
    % While EVO~\cite{GWPHKU:EVO}, ESVO~\cite{GWPHKU:ESVO}. ESVO2~\cite{ESVO_IMU,ESVO2}fail on all sequences.
    }
    \label{table:stereo_hku_Comparison_transposed}
    \resizebox{\columnwidth*2}{!}
    { 
    \begin{threeparttable}
    \begin{tabular}{c|c|ccccccccc|c} 
        \hline
        Methods 
        & Modality
        & agg\_translation
        & agg\_rotation 
        & agg\_flip 
        & agg\_walk 
        & hdr\_circle 
        & hdr\_slow 
        & hdr\_tran\_rota 
        & hdr\_agg 
        & dark\_normal 
        &Average\\
    \hline
    ORB-SLAM3~\cite{ORB-SLAM3}            & Stereo F+I         & 0.15  & 0.35  & 0.36  & \textit{failed} & 0.17  & 0.16  & 0.30  & 0.29  & \textit{failed} &0.25\\
    VINS-Fusion~\cite{GWPHKU:VINS-Fusion} & Stereo F+I       & 0.11  & 1.34  & 1.16  & \textit{failed} & 5.03  & 0.13  & 0.11  & 1.21  & 0.86 & 1.24\\
    EnVIO~\cite{envio}                    & Stereo F+I        &\textit{failed}        &2.12       &2.94       & 3.38         & 0.85      & 0.43      & 0.37      & 0.50      &\textit{failed}  & 1.51\\
    MSOC-S-IKF~\cite{MSOC-S-IKF}          & Stereo F+I  & \textit{failed} & 1.52 & \textit{failed} & \textit{failed} & \textit{failed} & \textit{failed}   & \textit{failed} & \textit{failed}  & \textit{failed} & 1.52\\
    DPVO~\cite{DPVO}                      & F                       & 0.07  & \textbf{0.04}  & 0.99  & 1.17            & 0.31  & 0.23  & 0.67  & 0.29  & \textit{failed} &0.47\\
    DBA-Fusion~\cite{DBA-Fusion}          & F+I               &  0.13     & 0.16      &  0.83     &0.37         &   0.18    & \textit{failed}      &  \textit{failed}    & 0.10      & 0.27 &0.29\\
    Kai et al.~\cite{tang2024monocular}   & E+I   &  0.21 & 0.28 & 0.81 & 0.35 & 0.71 & 0.43 & 0.50 & 0.27 &0.52 & 0.45\\
    PL-EVIO~\cite{GWPHKU:PL-EVIO} & E+F+I                & 0.07  & 0.23  & 0.39  & 0.42            & 0.14  & 0.13  & 0.10  & 0.14  & 1.35 &0.33\\
    EVI-SAM~\cite{GWPHKU:EVI-SAM} & E+F+I                & 0.17  & 0.24  & 0.32  & \textbf{0.26}            & \textbf{0.13}  & 0.11  & 0.11  & 0.10  & 0.85 &0.25\\
    % ESVO~\cite{GWPHKU:ESVO}               & Stereo E & \textit{failed} & \textit{failed} & \textit{failed} & \textit{failed} & \textit{failed} & \textit{failed}   & \textit{failed} & \textit{failed}  & \textit{failed} & ---\\
    ESIO~\cite{ESVIO} & Stereo E+I              & 0.55  & 0.78  & 3.17  & 1.30            & 0.46  & 0.31  & 0.91  & 1.41  & 0.35 &1.03\\
    ESVIO~\cite{ESVIO} & Stereo E+F+I           & 0.10  & 0.17  & 0.36  & 0.31            & 0.16  & 0.11  & 0.10  & 0.10  & 0.42 &0.20\\
    % EVO~\cite{GWPHKU:EVO}                 & E & \textit{failed} & \textit{failed} & \textit{failed} & \textit{failed} & \textit{failed} & \textit{failed}   & \textit{failed} & \textit{failed}  & \textit{failed} & ---\\
    DEVO~\cite{DEVO} & E                       & 0.06  & 0.05  & 0.71  & 0.90            & 0.39  & 0.08  & 0.08  & 0.26  & \textbf{0.06} &0.29\\
    \textbf{DEIO}                            & E+I        & \textbf{0.06}      &0.09       &\textbf{0.20}       &0.48                 & 0.14      & \textbf{0.07}      &0.09       & 0.06      &0.11 & \textbf{0.15}\\
    \hline
    \end{tabular}
    \end{threeparttable} 
    }
    \end{center}
    % \vspace{-1.6em}%调整表格与正文的距离
\end{table*}
%%%%%%%%%%%%%%%%%%***********************************************************************************************************************************************************%%%%%%%%%%%%%%%%%%

%%%%%%%%%%%%%%%%%%%%%%%%%%%%%%%%%%%%%%%%%%%%%%%%%%%%%%%%%%%%%%%%%%%%%%%%%%%%%%%%%%%%%%%%%%%%%%%%%%%%%%%%%%%%
\subsubsection*{\textbf{Stereo HKU-dataset}~\cite{ESVIO}} 

This dataset contains stereo event data at 60 Hz and stereo image frames at 30 Hz with a resolution of 346$\times$260, as well as IMU data at 1000 Hz.
It consists of handheld sequences including rapid motion and HDR scenarios.
In Table~\ref{table:stereo_hku_Comparison_transposed}, our method outperforms all previous works in terms of average positioning error.
Note that event-only VO methods, such as EVO~\cite{GWPHKU:EVO}, ESVO~\cite{GWPHKU:ESVO}, as well as stereo event and IMU-based methods like ESVO2~\cite{ESVO_IMU,ESVO2}, fail to perform successfully on any of the sequences in this dataset.
Moreover, DEIO beats DEVO and increases the average accuracy of the sequences up to 48\%.
Especially on \textit{agg\_walk} and \textit{agg\_flip}, DEVO encounters significant errors due to the limitations of relying solely on a monocular event camera. 
As shown in Fig.~\ref{figure:comparison_hku}(b), the trajectory estimated by DEIO closely aligns with the ground truth, whereas the baseline suffers from significant scale loss. 
It is important to note that the alignment of the estimated trajectories with the ground truth is computed using the publicly available trajectory evaluation tool~\cite{GWPHKU:evo_package}. 
We configure the tool to not align the scale of the estimated trajectories with the ground truth, ensuring an unbiased comparison of scale discrepancies. 
This underscores the importance of leveraging IMU to mitigate the degradation and scale ambiguity inherent in purely event-based VO.

%%%%%%%%%%%%%%%%%%***********************************************************************************************************************************************************%%%%%%%%%%%%%%%%%%
\begin{table*}[htb]
\renewcommand\arraystretch{1.2}
\begin{center}
\captionsetup{justification=justified}
\caption{
Accuracy comparison [MPE(\%)] of our DEIO with other image/event-based baselines in Vector dataset~\cite{GWPHKU:VECtor}.
The entire sequence of estimated poses is aligned with the ground truth trajectory.
}
\label{table:vector_Comparison}
\resizebox{2.0\columnwidth}{!}{
\begin{threeparttable}
\begin{tabular}{c|c|cccccccc|c} 
\hline  
\multirow{1}{*}{Methods} 
& \multirow{1}{*}{Modality} 
& \makecell{corner-\\slow} 
& \makecell{desk-\\normal} 
& \makecell{sofa-\\fast} 
& \makecell{mountain-\\fast} 
& \makecell{corridors-\\dolly} 
& \makecell{corridors-\\walk} 
& \makecell{units-\\dolly} 
& \makecell{units-\\scooter} 
& \makecell{average} \\ 
\hline
ORB-SLAM3~\cite{ORB-SLAM3} & Stereo F+I & 1.49 & 0.46 & 0.21 & 2.11 & 1.03 & 1.32 & 7.64 & 6.22 & 2.81 \\ 
VINS-Fusion~\cite{GWPHKU:VINS-Fusion} & Stereo F+I & 1.61 & 0.47 & 0.57 & \textit{failed} & 1.88 & 0.50 & 4.39 & 4.92 & 2.05 \\ 
DPVO~\cite{DPVO} & F & 0.30 & \textbf{0.09} & \textbf{0.07} & \textbf{0.11} & 0.56 & 0.54 & 1.52 & 1.67 & 0.61 \\ 
DBA-Fusion~\cite{DBA-Fusion} & F+I & 1.72 & 0.48 & 0.43 & \textit{failed} & 1.37 & 0.59 & 1.23 & 0.48 & 0.90 \\ 
EVO~\cite{GWPHKU:EVO} & E & 4.33 & \textit{failed} & \textit{failed} & \textit{failed} & \textit{failed} & \textit{failed} & \textit{failed} & \textit{failed} & 4.33\\ 
ESVO~\cite{GWPHKU:ESVO} & Stereo E & 4.83 & \textit{failed} & \textit{failed} & \textit{failed} & \textit{failed} & \textit{failed} & \textit{failed}  &\textit{failed}  & 4.83  \\ 
Ultimate-SLAM~\cite{GWPHKU:Ultimate-SLAM} & E+F+I & 4.83 & 2.24 & 2.54 & 4.13 & \textit{failed} & \textit{failed} & \textit{failed} & \textit{failed}  &  3.44\\ 
PL-EVIO~\cite{GWPHKU:PL-EVIO} & E+F+I & 2.10 & 3.66 & 0.17 & 0.13 & 1.58 & 0.92 & 5.84 & 5.00 & 2.92 \\ 
ESVIO~\cite{ESVIO} & Stereo E+F+I & 1.49 & 0.61 & 0.17 & 0.16 & 1.13 & 0.43 & 3.43 & 2.85 & 1.41 \\ 
EVI-SAM~\cite{GWPHKU:EVI-SAM} & E+F+I & 2.50 & 1.45 & 0.98 & 0.38 & 1.58 & 1.27 & 0.59 & 0.83 & 1.32 \\ 
DEVO~\cite{DEVO} & E & 0.59 & 0.11 & 0.38 & 0.37 & \textbf{0.51} & 1.04 & 0.48 & 0.88 & 0.55 \\ 
\textbf{DEIO} & E+I & \textbf{0.50} & 0.13 & 0.44 & 0.24 & 0.78 & 0.74 & \textbf{0.35} & \textbf{0.35} & \textbf{0.44} \\ 
\hline
\end{tabular}
\end{threeparttable} 
}
\end{center}
 % \vspace{-1.5em}%调整表格与正文的距离
\end{table*}

%%%%%%%%%%%%%%%%%%***********************************************************************************************************************************************************%%%%%%%%%%%%%%%%%%

%%%%%%%%%%%%%%%%%%%%%%%%%%%%%%%%%%%%%%%%%%%%%%%%%%%%%%%%%%%%%%%%%%%%%%%%%%%%%%%%%%%%%%%%%%%%%%%%%%%%%%%%%%%%
\subsubsection*{\textbf{VECtor}~\cite{GWPHKU:VECtor}} 

This dataset consists of a hardware-synchronized sensor suite that includes stereo event cameras (640$\times$480), stereo standard cameras (1224$\times$1024), and IMU.
It covers the full spectrum of 6 DoF motion dynamics, environment complexities, and illumination conditions for both small and large-scale scenarios.
As presented in Table~\ref{table:vector_Comparison}, our proposed DEIO achieves remarkable results on average.
It surpasses all image-based baselines with high-quality frames and even outperforms Ultimate-SLAM, PL-EVIO, ESVIO, and EVI-SAM on over 75\% of the sequences, which utilize event, image, and IMU.
DEIO also outperforms DEVO on average in large-scale sequences, thanks to the complementary integration of the event and IMU sensors, while other monocular visual-only methods struggle with scale ambiguity and drift.

%%%%%%%%%%%%%%%%%%%%%%%%%%%%%%%%%%%%%%%%%%%%%%%%%%%%%%%%%%%%%%%%%%%%%%%%%%%%%%%%%%%%%%%%%%%%%%%%%%%%%%%%%%%%
\subsubsection*{\textbf{TUM-VIE}~\cite{TUM-VIE}} 

This dataset is recorded with stereo high-resolution event cameras (1280$\times$720) mounted on a helmet, capturing footage from an egocentric viewpoint. 
We evaluate our method on sequences recorded in a motion capture room that featured various 6-DOF camera movements.
The results in Table~\ref{table:tum-vie} demonstrate that DEIO outperforms all other methods on four out of five sequences, despite DH-PTAM~\cite{DH-PTAM} utilizing four cameras of the setup (stereo events and stereo images).
It is worth noting that, due to the relatively low complexity of the selected sequence, the learning-based event data association method is already highly effective in these scenarios. 
For instance, the performance of the purely event-based VO (DEVO) achieves an accuracy of 1.2 cm. 
While the introduction of IMU in DEIO still brings an enhancement by the additional constraints from the IMU measurements.

%%%%%%%%%%%%%%%%%%***********************************************************************************************************************************************************%%%%%%%%%%%%%%%%%%
\begin{table}[htb]
         % \vspace{-1.0em}%调整表格与正文的距离
        \renewcommand\arraystretch{1.2}
        \begin{center}
        \captionsetup{justification=justified}
        \caption{
        Accuracy comparison [ATE/RMSE (cm)] of our DEIO with other event-based baselines in TUM-VIE dataset~\cite{TUM-VIE}.
        The entire sequence of estimated poses is aligned with the ground truth trajectory.
        The baseline results (EVO, ESVO, and ES-PTAM) are taken from~\cite{ES-PTAM}, while DH-PTAM and Ultimate-SLAM are sourced from~\cite{DH-PTAM}, DEVO is from~\cite{DEVO}, and ESVIO\_AA, ESVO2 are from~\cite{ESVO2}.
        }
        \label{table:tum-vie}
        \resizebox{\columnwidth}{!}
        { 
        \begin{threeparttable}
        \begin{tabular}{c|c|ccccc|c} 
        \hline  
     \multirow{2}{*}{Methods} & \multirow{2}{*}{Modality} & \multicolumn{5}{c|}{mocap-} & \multirow{2}{*}{Average} \\
    \cline{3-7}
    ~& ~& 1d-trans & 3d-trans & 6dof & desk & desk2 & \\
\hline
EVO~\cite{GWPHKU:EVO}                     & E          & 7.5 &12.5 &85.5 &54.1 &75.2 & 47.0 \\
ESVO~\cite{GWPHKU:ESVO}                   & Stereo E   & 12.3 & 17.2 & 13.0 & 12.4 & 4.6 &  11.9\\
ESVIO\_AA~\cite{ESVO_IMU}             & Stereo E+I   & 3.9 & 18.9 & \textit{failed} & 9.00 & 9.5 &10.3  \\
ESVO2~\cite{ESVO2}                   & Stereo E+I   & 3.3 & 7.3 & 3.2 & 6.2 & 4.0 & 4.8 \\
ES-PTAM~\cite{ES-PTAM}                    & Stereo E   & 1.05 & 8.53 & 10.25 & 2.5 & 7.2 & 5.9 \\
DH-PTAM~\cite{DH-PTAM}                    & Stereo E+F &10.3 & \textbf{0.7} & 2.4 & 1.6 & 1.5 & 3.9 \\
Ultimate-SLAM~\cite{GWPHKU:Ultimate-SLAM} & E+F+I      & 3.9 & 4.7 & 35.3 &19.5  &34.1 &  19.5\\
DEVO~\cite{DEVO}                          & E & 0.5 & 1.1 & 1.6 & 1.7 & 1.0 &  1.2\\
% DVIO$^{\dag}$                             & F+I    &6.7 & 16.9 & 9.8 & 14.1 & 14.0 & 12.5 \\
\textbf{DEIO}                             & E+I  & \textbf{0.4} & 1.1 & \textbf{1.4} & \textbf{1.4} & \textbf{0.7} & \textbf{1.0} \\
\hline
        \end{tabular}
        \end{threeparttable} 
        }
        \end{center}
        % \vspace{-2.0em}
\end{table}
%%%%%%%%%%%%%%%%%%***********************************************************************************************************************************************************%%%%%%%%%%%%%%%%%%

%%%%%%%%%%%%%%%%%%%%%%%%%%%%%%%%%%%%%%%%%%%%%%%%%%%%%%%%%%%%%%%%%%%%%%%%%%%%%%%%%%%%%%%%%%%%%%%%%%%%%%%%%%%%
\subsubsection*{\textbf{EDS}~\cite{EDS}} 

\begin{figure*}[htb]  %%(h 此处（here） t 页顶（top）b 页底（bottom） p 独立一页（page）)
        \centering        
        \captionsetup{justification=justified}%图题对齐
        \includegraphics[width=2.0\columnwidth]{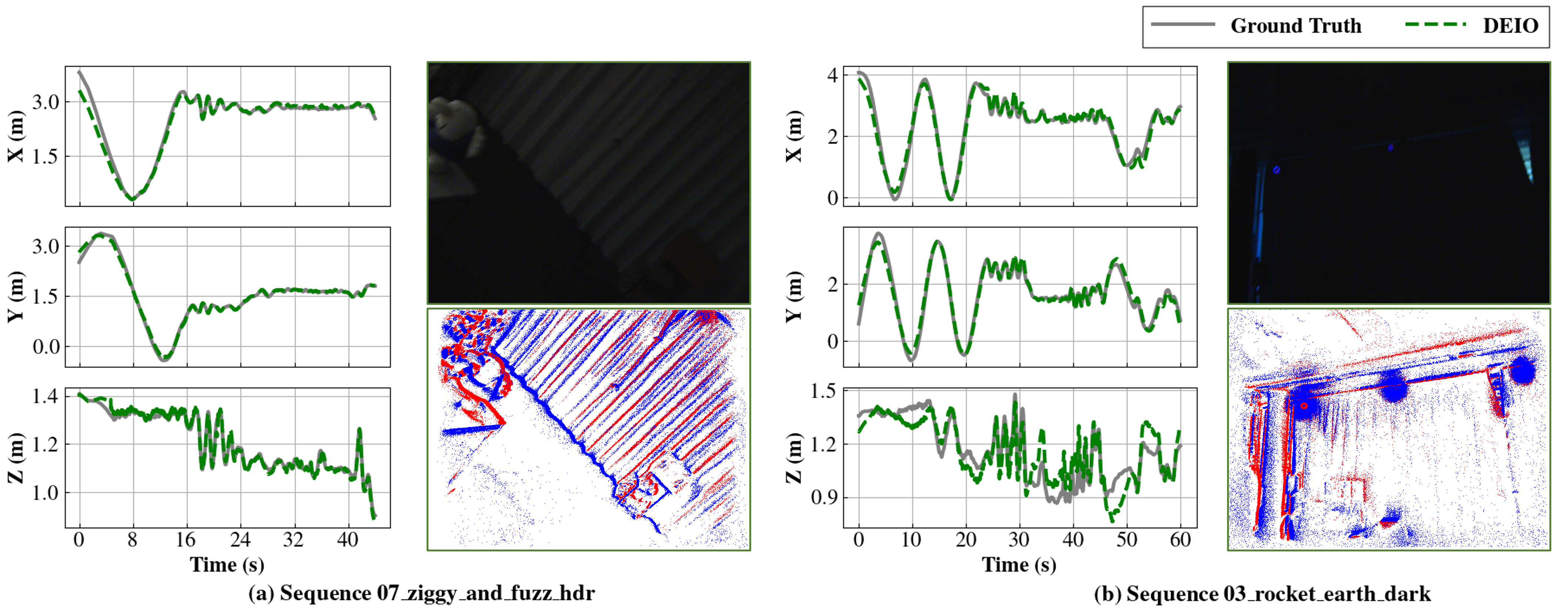}
        \caption{
        The estimated trajectories of our DEIO against the GT in the sequence of ziggy\_hdr and rocket\_dark from the EDS~\cite{EDS} dataset. 
        The image view (visualization-only) demonstrates the lack of perceptible information under low-light conditions, while the event view, though perceptible, remains susceptible to interference from the infrared light of the motion capture system. 
        Thanks to our robust learning-based event data association, the trajectories estimated by DEIO align remarkably closely with the GT.
        }  %设置图片的一个编号以及为图片添加标题
        \label{figure:deio_eds}
\end{figure*}

This dataset provides synchronized monocular event (640$\times$480), image (640$\times$480), and IMU data from handheld devices with ground truth poses.
To the best of our knowledge, this work presents the first results for event-inertial pose estimation using the EDS dataset.
As shown in Table~\ref{table:eds_Comparison}, our DEIO outperforms the image-based baselines, including ORB-SLAM3, DPVO, and DBA-Fusion.
Moreover, DEIO achieves an average improvement of 30\% over DEVO and demonstrates performance comparable to RAMP-VO, a learning-based VO system that leverages both event and image modalities.
In the case of DBA-Fusion~\cite{DBA-Fusion}, even when operating under a learning-based VIO framework, DEIO demonstrates superior performance, showing the distinct advantages of event cameras in challenging scenarios.
For instance, the HDR characteristics in the sequence \textit{ziggy\_hdr} cause severe image degradation (as illustrated in Fig.~\ref{figure:deio_eds}).
In contrast, DEIO leverages the rich perceptual information provided by the event camera, achieving an 80\% improvement in accuracy over DBA-Fusion~\cite{DBA-Fusion} in \textit{ziggy\_hdr}.

%%%%%%%%%%%%%%%%%%***********************************************************************************************************************************************************%%%%%%%%%%%%%%%%%%
\begin{table*}[htb]
    \renewcommand\arraystretch{1.2}
    \begin{center}
    \captionsetup{justification=justified}
    \caption{Accuracy comparison [ATE/RMSE (cm)] of our DEIO with other image/event-based baselines in EDS dataset~\cite{EDS}.
    The entire sequence of estimated poses is aligned with the ground truth trajectory.
    The baseline results (ORB-SLAM3, DPVO, and DEVO) are taken from~\cite{DEVO}, while RAMP-VO is sourced from~\cite{rampvo}.
        }
    \label{table:eds_Comparison}
    \resizebox{\columnwidth*2}{!}
    { 
    \begin{threeparttable}
    \begin{tabular}{c|c|ccccccccc|c} 
        \hline
        Methods 
        & Modality
        & peanuts\_dark
        & peanuts\_light
        & peanuts\_run 
        & rocket\_dark
        & rocket\_light 
        & ziggy
        & ziggy\_hdr 
        & ziggy\_flying
        & all\_chars 
        &Average\\
    \hline
    ORB-SLAM3~\cite{ORB-SLAM3}            & Stereo F+I      & 6.15  & 27.26  & 16.83 &  10.12 & 32.53 & 26.92 & 81.98 & 20.57 & 21.37 & 27.08\\
    DPVO~\cite{DPVO}                      & F               & 1.26  & 12.99  & 25.48 &  27.41 & 63.11 & 14.86 & 66.17 & 10.85 & 95.87 & 35.33\\
    DBA-Fusion~\cite{DBA-Fusion}          & F+I             & 7.26  & 149.36 & 134.92&  114.24& 117.09& 173.50& 140.51& 11.81 & 126.36& 108.33\\
    DEVO~\cite{DEVO}                      & E               & 4.78  & 21.07 & 38.10 &  8.78  & 59.83 & 11.84 & \textbf{22.82} & 10.92 & \textbf{10.76} & 21.00\\
    RAMP-VO~\cite{rampvo}       & E+F             & \textbf{1.20}  & \textbf{9.03}  & \textbf{13.19} &  \textbf{7.20}  & 17.53 & 19.05 & 28.78 & 6.35  & 28.61 & \textbf{14.55}\\
    \textbf{DEIO}                         & E+I             & 1.77  & 16.27 & 19.96 &  8.91  & \textbf{15.41} & \textbf{10.39} & 23.82  & \textbf{3.84}   & 31.55 & \textbf{14.66}\\
    \hline
    \end{tabular}
    \end{threeparttable} 
    }
    \end{center}
     % \vspace{-1.5em}%调整表格与正文的距离
\end{table*}
%%%%%%%%%%%%%%%%%%***********************************************************************************************************************************************************%%%%%%%%%%%%%%%%%%

%%%%%%%%%%%%%%%%%%***********************************************************************************************************************************************************%%%%%%%%%%%%%%%%%%
%%%%%%%%%%%%%%%%%%***********************************************************************************************************************************************************%%%%%%%%%%%%%%%%%%

%%%%%%%%%%%%%%%%%%%%%%%%%%%%%%%%%%%%%%%%%%%%%%%%%%%%%%%%%%%%%%%%%%%%%%%%%%%%%%%%%%%%%%%%%%%%%%%%%%%%%%%%%%%%
\subsubsection*{\textbf{MVSEC}~\cite{GWPHKU:MVSEC}} 

\begin{figure*}[htb]  %%(h 此处（here） t 页顶（top）b 页底（bottom） p 独立一页（page）)
        \centering        
        \captionsetup{justification=justified}%图题对齐
        \includegraphics[width=2.0\columnwidth]{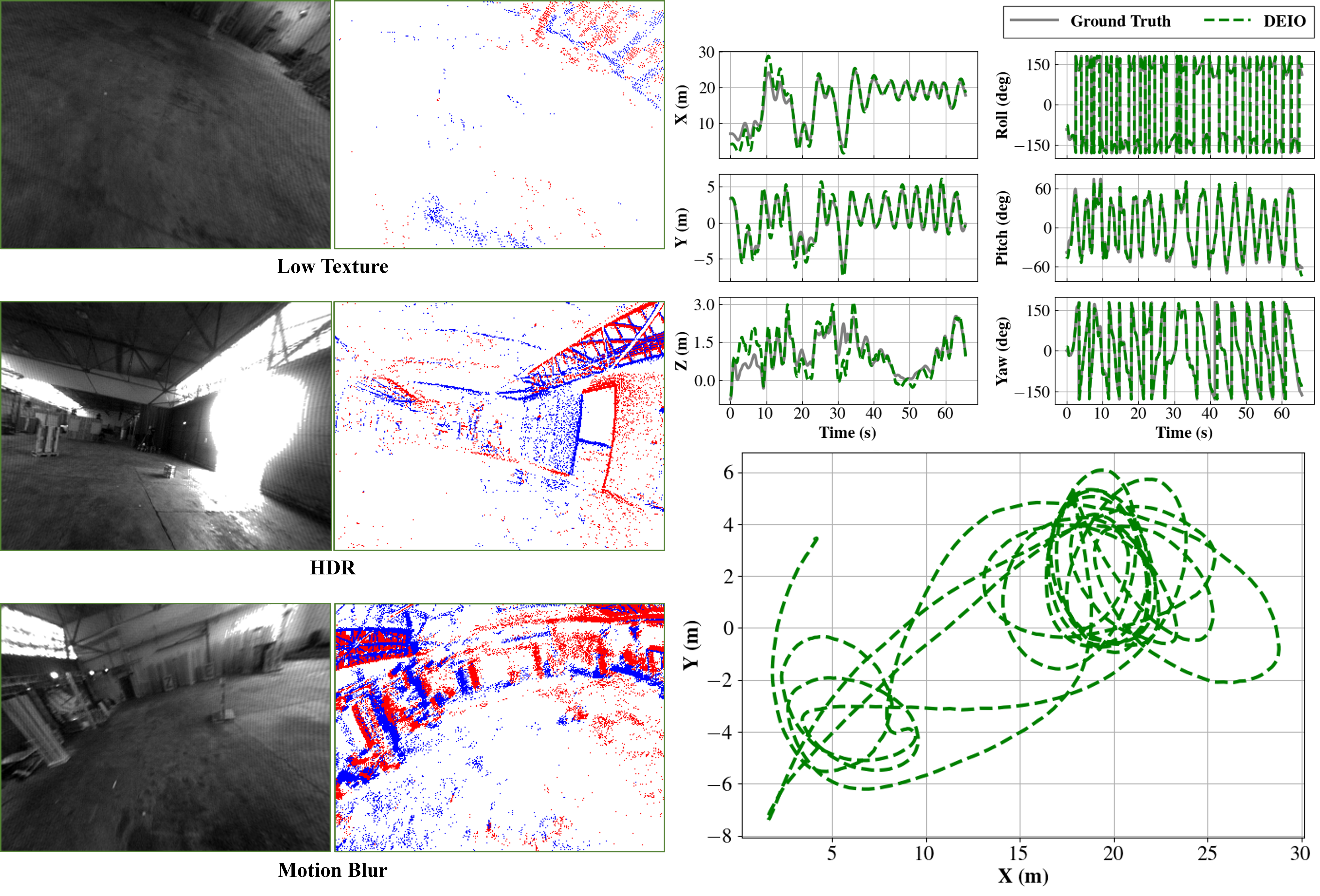}
        \caption{
        The estimated trajectories (X, Y, Z, Roll, Pitch, Yaw) of our DEIO against the GT in the sequence of indoor\_forward\_7 from the UZH-FPV~\cite{GWPHKU:FPV} dataset. 
        The image view (visualization-only) demonstrates the condition under low texture, HDR, and motion blur.
        }  %设置图片的一个编号以及为图片添加标题
        \label{figure:deio_FPV}
\end{figure*}

This dataset is collected by stereo DAVIS346.
Specifically, we select the \textit{indoor\_flying} sequence, which features arbitrary 6-DOF motion from a drone flying in a motion capture room. 
The forward and backward movements of the drone, along with pauses at the start and end points that produce almost no events, make these sequences particularly challenging.
Additionally, the sparse events generated far from the camera pose an additional challenge for event-based SLAM.
In Table~\ref{table:mvsec_Comparison}, DEIO surpasses the event-based baseline, especially for the \textit{Flying\_4}, where it attains an RMSE of 40\% lower than DEVO.
Although ESVIO~\cite{ESVIO} employs a setup integrating stereo images, stereo events, and IMU data, DEIO, which relies solely on monocular event data and IMU, demonstrates an average accuracy improvement of over 80\% compared to ESVIO.
As for the learning-based VIO method (DBA-Fusion) that relies on dense data association, it failed on three out of the four sequences.
This indicates that even though the deep learning methods can provide strong data association capabilities, the degradation of images in challenging environments limits their performance compared to the event modality. 
Another example is illustrated in Fig.~\ref{figure:deio_FPV}.

%%%%%%%%%%%%%%%%%%***********************************************************************************************************************************************************%%%%%%%%%%%%%%%%%%
\begin{table}[htb]
         % \vspace{-1.0em}%调整表格与正文的距离
        \renewcommand\arraystretch{1.2}
        \begin{center}
        \captionsetup{justification=justified}%图题对齐
        \caption{Accuracy comparison [MPE (\%)] of our DEIO with other image/event-based baselines in MVSEC dataset~\cite{GWPHKU:MVSEC}.
        The entire sequence of estimated poses is aligned with the ground truth trajectory.
        The DEVO result is taken from~\cite{DEVO}.
        }
        \label{table:mvsec_Comparison}
        \resizebox{\columnwidth}{!}
        { 
        \begin{threeparttable}
        \begin{tabular}{c|c|cccc|c} 
        \hline  
    \multirow{1}*{Methods} 
    & \multirow{1}*{Modality}
    & Flying\_1 & Flying\_2 & Flying\_3 & Flying\_4 & Average\\
\hline
ORB-SLAM3~\cite{ORB-SLAM3}                & Stereo F+I & 5.31  & 5.65  & 2.90 & 6.99 & 5.21  \\
VINS-Fusion~\cite{GWPHKU:VINS-Fusion}     & Stereo F+I & 1.50  & 6.98  & 0.73  & 3.62 & 3.21 \\
EVO~\cite{GWPHKU:EVO}                     & E          & 5.09  & \textit{failed} & 2.58 & \textit{failed} & 3.84 \\
ESVO~\cite{GWPHKU:ESVO}                   & Stereo E   & 4.00  & 3.66 & 1.71 & \textit{failed} & 3.12 \\
Ultimate-SLAM~\cite{GWPHKU:Ultimate-SLAM} & E+F+I & \textit{failed} & \textit{failed} & \textit{failed} & 2.77  & 2.77\\
PL-EVIO~\cite{GWPHKU:PL-EVIO}             & E+F+I      & 1.35  & 1.00  & 0.64  & 5.31 & 2.08\\
ESVIO~\cite{ESVIO}                        & Stereo E+F+I & 0.94 & 1.00 & 0.47 & 5.55 & 1.99\\
DBA-Fusion~\cite{DBA-Fusion}              & F+I & 2.20 & \textit{failed} & \textit{failed} & \textit{failed} & 2.20 \\
DEVO~\cite{DEVO}                          & E & 0.26  & 0.32 & 0.19  & 1.08 & 0.46 \\
\textbf{DEIO}                                        & E+I & \textbf{0.24} & \textbf{0.21} & \textbf{0.12} & \textbf{0.78} & \textbf{0.34}\\
\hline
\end{tabular}
\end{threeparttable} 
}
\end{center}
\end{table}
%%%%%%%%%%%%%%%%%%***********************************************************************************************************************************************************%%%%%%%%%%%%%%%%%%

%%%%%%%%%%%%%%%%%%***********************************************************************************************************************************************************%%%%%%%%%%%%%%%%%%
%%%%%%%%%%%%%%%%%%***********************************************************************************************************************************************************%%%%%%%%%%%%%%%%%%

%%%%%%%%%%%%%%%%%%***********************************************************************************************************************************************************%%%%%%%%%%%%%%%%%%
\begin{table}[htb] 
    \renewcommand\arraystretch{1.2}
    \begin{center}
    \captionsetup{justification=justified}%图题对齐
    \caption{Accuracy comparison [MPE (\%)] of our DEIO with other image/event-based baselines in UZH-FPV dataset~\cite{GWPHKU:FPV}.
    The entire sequence of estimated poses is aligned with the ground truth trajectory.
    The baseline results (DPVO, DEVO) are taken from~\cite{DEVO}.
    }
    \label{table:UZH-FPV_Comparison}
    \resizebox{\columnwidth}{!}
    { 
    \begin{threeparttable}
    \setlength{\tabcolsep}{1.0mm}
    \begin{tabular}{c|c|cccccc|c} 
    \hline
    \multirow{2}{*}{Methods} & \multirow{2}{*}{Modality} & \multicolumn{6}{c|}{Indoor\_forward} & \multirow{2}{*}{Average} \\
    \cline{3-8}
    ~& ~& 3 & 5 & 6 & 7 & 9 & 10 & \\
    \hline
    VINS-Fusion~\cite{GWPHKU:VINS-Fusion} & Stereo F+I & 0.84 & \textit{failed} & 1.45 & 0.61 & 2.87 & 4.48 & 2.45 \\
    ORB-SLAM3~\cite{ORB-SLAM3} & Stereo F+I & 0.55 & 1.19 & \textit{failed} & 0.36 & 0.77 & 1.02 & 0.78 \\
    EVO~\cite{GWPHKU:EVO} & E & \textit{failed} & \textit{failed} & \textit{failed} & \textit{failed} & \textit{failed} & \textit{failed} & --- \\
    DPVO~\cite{DPVO} & F & \textit{failed} & \textit{failed} & \textit{failed} & \textit{failed} & \textit{failed} & \textit{failed} & --- \\
    VINS-MONO~\cite{GWPHKU:VINS-MONO} & F+I & 0.65 & 1.07 & \textbf{0.25} & 0.37 & 0.51 & 0.92 & 0.63 \\
    DBA-Fusion~\cite{DBA-Fusion} & F+I &\textit{failed}   &\textit{failed}  &\textit{failed}  &\textit{failed}  & \textit{failed} &\textit{failed}  &  --- \\
    Ultimate SLAM~\cite{GWPHKU:Ultimate-SLAM} & E+F+I & \textit{failed} & \textit{failed} & \textit{failed} & \textit{failed} & \textit{failed} & \textit{failed} & --- \\
    PL-EVIO~\cite{GWPHKU:PL-EVIO} & E+F+I & 0.38 & 0.90 & 0.30 & 0.55 & \textbf{0.44} & 1.06 & 0.60 \\
    DEVO~\cite{DEVO} & E & \textbf{0.37} & 0.40 & 0.31 & 0.50 & 0.61 & \textbf{0.52} & 0.45 \\
    \textbf{DEIO} & E+I & 0.39  & \textbf{0.36} & 0.33 & \textbf{0.32} & 0.59 & 0.55 & \textbf{0.42} \\
    \hline    
    \end{tabular} 
    \end{threeparttable} 
    }
    \end{center}
    % \vspace{-1.5em}
\end{table}
%%%%%%%%%%%%%%%%%%***********************************************************************************************************************************************************%%%%%%%%%%%%%%%%%%

%%%%%%%%%%%%%%%%%%%%%%%%%%%%%%%%%%%%%%%%%%%%%%%%%%%%%%%%%%%%%%%%%%%%%%%%%%%%%%%%%%%%%%%%%%%%%%%%%%%%%%%%%%%%
\subsubsection*{\textbf{UZH-FPV}~\cite{GWPHKU:FPV}} 

This dataset, captured using the monocular DAVIS346, consists of high-speed trajectories, including fast laps around a racetrack with drone racing gates and free-form paths around various obstacles.
As shown in Table~\ref{table:UZH-FPV_Comparison} and Fig.~\ref{figure:deio_FPV}, this dataset poses significant challenges for existing methods, with even advanced learning-based VO approaches like DPVO failing to maintain reliable tracking across all sequences.
Additionally, incorporating IMU measurements does not resolve these challenges, such as learning-based VIO methods (DBA-Fusion), which also fail to complete any sequences.
This is due to the motion blur caused by rapid movement, which makes it difficult to effectively establish data association for the image sensor, even if these methods are equipped with a powerful learning network.
In contrast, our DEIO achieves higher average performance than all baseline methods across all sequences, demonstrating greater resilience to the fast flight conditions.
The modest improvement from DEVO to DEIO can be attributed to significant IMU bias induced by aggressive motion, which limits the potential benefits of IMU-aided optimization.
Nevertheless, our DEIO still successfully handles complex flight scenarios that involve challenging maneuvers, such as back-and-forth motion, abrupt directional changes, and loops.

%%%%%%%%%%%%%%%%%%***********************************************************************************************************************************************************%%%%%%%%%%%%%%%%%%
\begin{figure*}[htb]  %%(h 此处（here） t 页顶（top）b 页底（bottom） p 独立一页（page）)
        \centering     
        \captionsetup{justification=justified}%图题对齐
        \includegraphics[width=1.8\columnwidth]{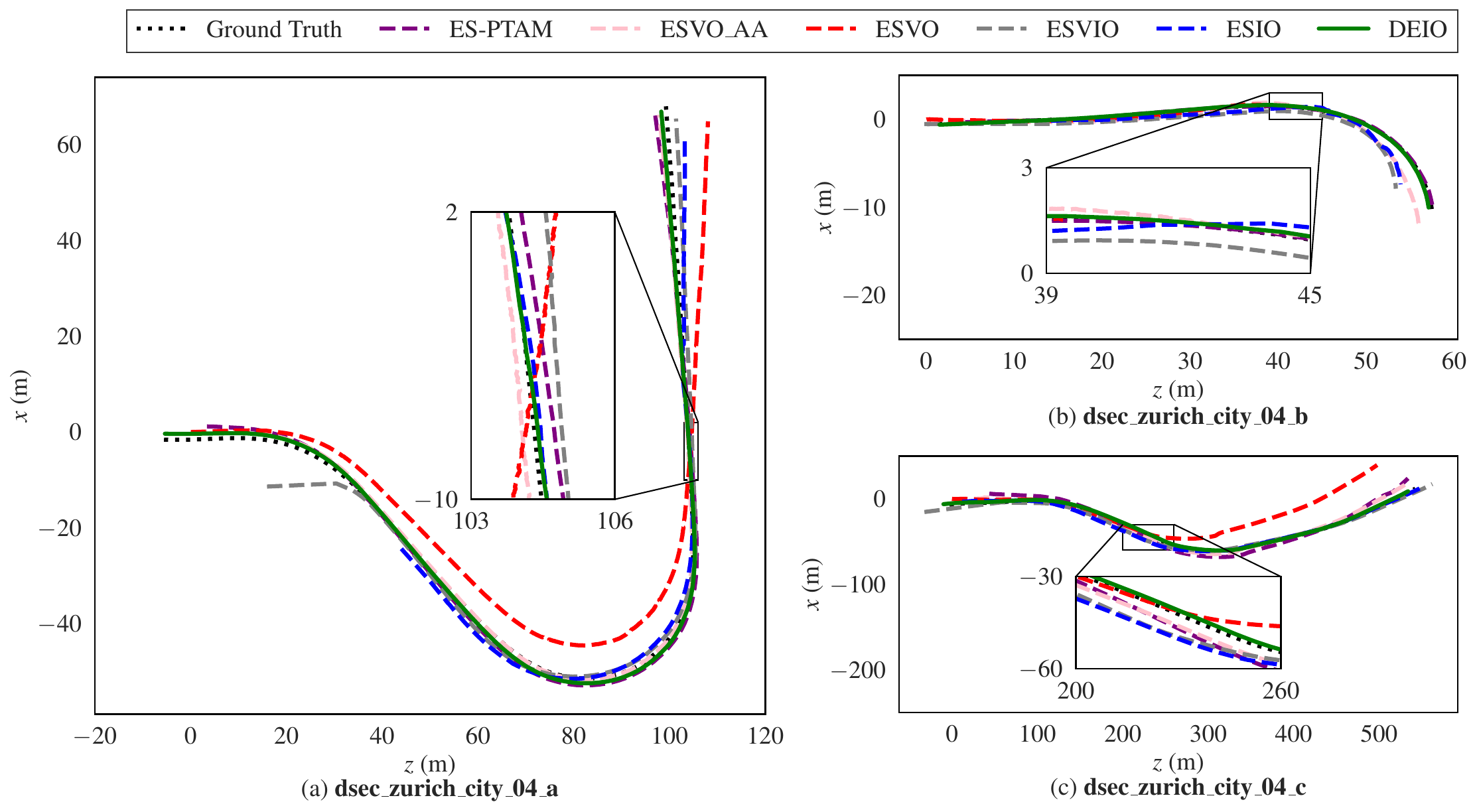}  
        \caption{
         Visualization of estimated trajectories in the DSEC dataset~\cite{GWPHKU:DSEC}.
         Despite using a monocular setup, our DEIO results align more closely with the ground truth compared to other methods, which utilize a stereo setup.
        }  %设置图片的一个编号以及为图片添加标题
        \label{figure:desc_result}
        % \vspace{-1.6em}%调整表格与正文的距离
\end{figure*}%
%%%%%%%%%%%%%%%%%%***********************************************************************************************************************************************************%%%%%%%%%%%%%%%%%%

%%%%%%%%%%%%%%%%%%***********************************************************************************************************************************************************%%%%%%%%%%%%%%%%%%
\begin{figure*}[htb]  %%(h 此处（here） t 页顶（top）b 页底（bottom） p 独立一页（page）)
        \centering     
        \captionsetup{justification=justified}%图题对齐
        \includegraphics[width=2.0\columnwidth]{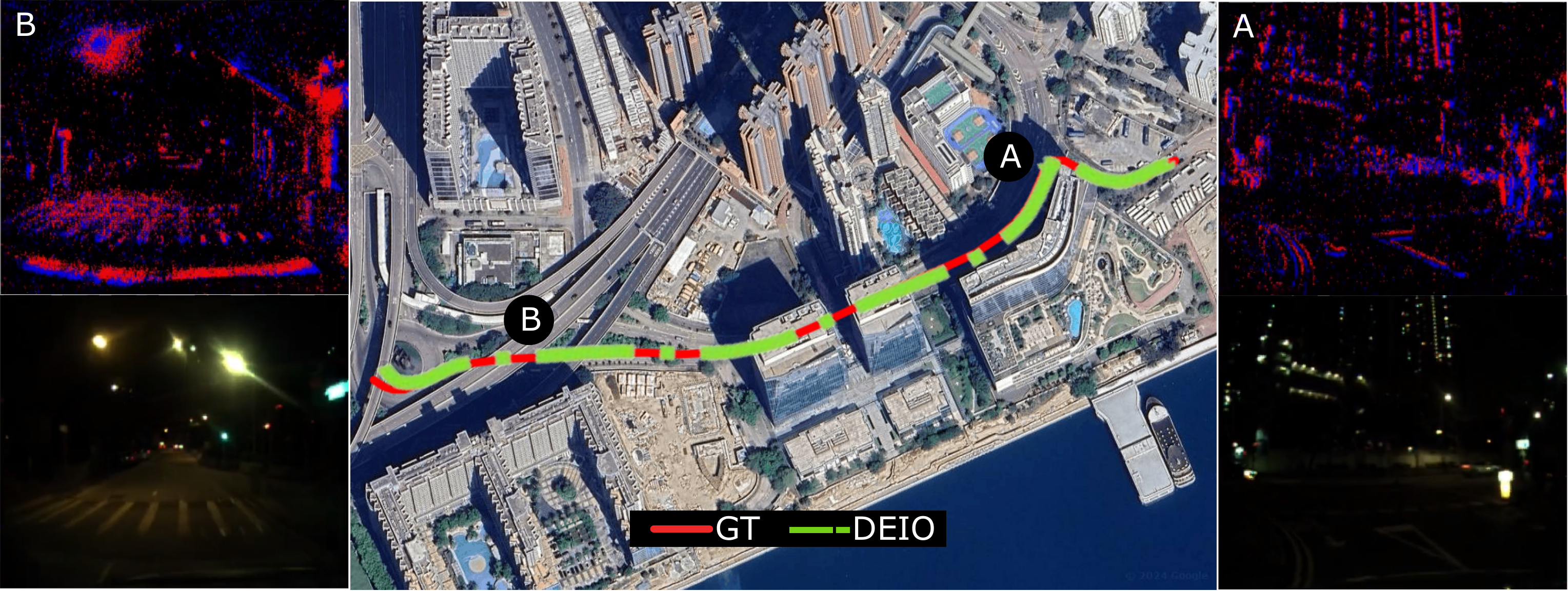}  
        \caption{
        % (a) The driving platform~\cite{GWPHKU:ECMD}.
        % (b)
        The estimated trajectory of our DEIO in the night driving scenarios~\cite{GWPHKU:ECMD} and its comparison against the GNSS-INS-RTK as ground truth.
        The image view is for visualization only.
        }  %设置图片的一个编号以及为图片添加标题
        \label{figure:ecmd_testing}
        % \vspace{-1.6em}%调整表格与正文的距离
\end{figure*}%
%%%%%%%%%%%%%%%%%%***********************************************************************************************************************************************************%%%%%%%%%%%%%%%%%%

%%%%%%%%%%%%%%%%%%%%%%%%%%%%%%%%%%%%%%%%%%%%%%%%%%%%%%%%%%%%%%%%%%%%%%%%%%%%%%%%%%%%%%%%%%%%%%%%%%%%%%%%%%%%
\subsubsection*{\textbf{DSEC}~\cite{GWPHKU:DSEC}}

This dataset is collected with stereo event cameras (640$\times$480) mounted on a car driving through the streets of Switzerland. 
Since the DSEC dataset lacks ground truth trajectories, we use the odometry trajectories provided by~\cite{ESVO_IMU}, which are obtained through a LiDAR-IMU-based method, as ground truth.
As presented in Table~\ref{table:DSEC_Comparison}, our DEIO outperforms the stereo event methods (ES-PTAM, ESVO, ESVIO\_AA, and ESIO) by large margins on all sequences (at least 66.7\% lower RMSE).
Fig.~\ref{figure:desc_result} presents a qualitative comparison of the estimated trajectories of our DEIO and these stereo event-based baselines.
The results from our DEIO align more closely with the ground truth, despite using a monocular setup, while the other methods employ a stereo setup.
This demonstrates that DEIO can achieve comparable scale estimation to these stereo setups while providing superior state estimation results.
Furthermore, as can be seen from the results, the purely event-based odometry (ESVO) exhibits the most significant drift, highlighting the importance of complementarity between events and IMU sensors.

%%%%%%%%%%%%%%%%%%***********************************************************************************************************************************************************%%%%%%%%%%%%%%%%%%
\begin{table}[htb] 
    \renewcommand\arraystretch{1.2}
    \begin{center}
    \captionsetup{justification=justified}%图题对齐
    \caption{Accuracy comparison [ATE/RMSE (cm)] of our DEIO with stereo event-based methods in DSEC dataset~\cite{GWPHKU:DSEC}.
    The entire sequence of estimated poses is aligned with the ground truth trajectory.
    The baseline results (ESVO and ESVIO\_AA) are taken from~\cite{ESVO_IMU}, and ES-PTAM is sourced from~\cite{ES-PTAM}.
    }
    \label{table:DSEC_Comparison}
    \resizebox{\columnwidth}{!}
    { 
    \begin{threeparttable}
    \setlength{\tabcolsep}{1.0mm}
    \begin{tabular}{c|c|ccccc|c} 
    \hline
    \multirow{2}{*}{Methods} & \multirow{2}{*}{Modality} & \multicolumn{5}{c|}{dsec\_zurich\_city\_04}  & \multirow{2}{*}{Average} \\
    \cline{3-7}
    ~& ~& a & b & c & d & e & \\
    \hline
     ESVO~\cite{GWPHKU:ESVO}  & Stereo E    & 371.1  & 116.6  & 1357.1  & 2676.6  & 794.9  & 1032.9 \\
     ESVIO\_AA~\cite{ESVO_IMU} & Stereo E+I  & 105.0  & 66.7   & 637.9   & 699.8   & 130.3  & 527.9 \\
     ES-PTAM~\cite{ES-PTAM}   & Stereo E    & 131.62 & \textbf{29.02}  & 1184.37 & 1053.87 & \textbf{75.9}   &494.9 \\
     ESIO~\cite{ESVIO}        & Stereo E+I  & 543.5  & 295.1  & 896.2   & 2977.0  &2326.4   & 1587.8 \\
 ESVIO~\cite{ESVIO}       & Stereo E+F+I    & 371.2  & 445.8  & 1892.7  & 921.7  &352.0   & 596.9\\
     \textbf{DEIO}            & E+I & \textbf{80.6} & 35.4 & \textbf{413.8} & \textbf{207.6} & 86.1  & \textbf{164.5}  \\
    \hline    
    \end{tabular}
    \end{threeparttable} 
    }
    \end{center}
    % \vspace{-1.5em}%调整表格与正文的距离
\end{table}
%%%%%%%%%%%%%%%%%%***********************************************************************************************************************************************************%%%%%%%%%%%%%%%%%%

%%%%%%%%%%%%%%%%%%%%%%%%%%%%%%%%%%%%%%%%%%%%%%%%%%%%%%%%%%%%%%%%%%%%%%%%%%%%%%%%%%%%%%%%%%%%%%%%%%%%%%%%%%%%
\subsection{Test on Night Driving Scenarios}
\label{section:driving scenarios}
Driving scenarios pose challenges for event-based state estimation, especially at nighttime, where rampant flickering light (e.g., from LED signs) generates an overwhelming number of noisy events.
Additionally, the movement of vehicles, such as sharp turns, sudden stops, and other abrupt movements, can further complicate the event-based estimator.
In this section, we select the \textit{Dense\_street\_night\_easy\_a} sequences of the ECMD dataset~\cite{GWPHKU:ECMD}, which feature numerous flashing lights from vehicles, street signs, buildings, and moving vehicles, making event-based SLAM more difficult.
This dataset is recorded with two pairs of stereo event cameras (640$\times$480 and 346$\times$260) on a car driven through various road conditions such as streets, highways, roads, and tunnels in Hong Kong.
Our DEIO runs on the event from the DAVIS346 and the IMU sensor, while the image frame output of the DAVIS346 is only used for illustration purposes.
Fig.~\ref{figure:ecmd_testing} shows a small drift with a 4.7 m error of our estimated trajectory on the 620 m drive.
To the best of our knowledge, we present the first results on pose tracking for night driving scenarios using event and IMU odometry. 
The earliest attempt in this area is ESVIO~\cite{ESVIO, GWPHKU:ECMD}, which utilizes a combination of stereo events, stereo images, and IMU data, whereas DEIO operates with a monocular setup.

%%%%%%%%%%%%%%%%%%***********************************************************************************************************************************************************%%%%%%%%%%%%%%%%%%
\begin{figure*}[htb]  %%(h 此处（here） t 页顶（top）b 页底（bottom） p 独立一页（page）)
        \subfigtopskip=0pt %设置子图与上面正文或别的内容的距离
        \subfigbottomskip=0pt %设置第二行子图与第一行子图的距离，即下面的头与上面的脚的距离
        \subfigcapskip=-3pt %设置子图与子标题之间的距离
        \centering
        \begin{minipage}[t]{1.0\columnwidth}
                \centering
                \subfigure[ ]{
                \includegraphics[width=\columnwidth]{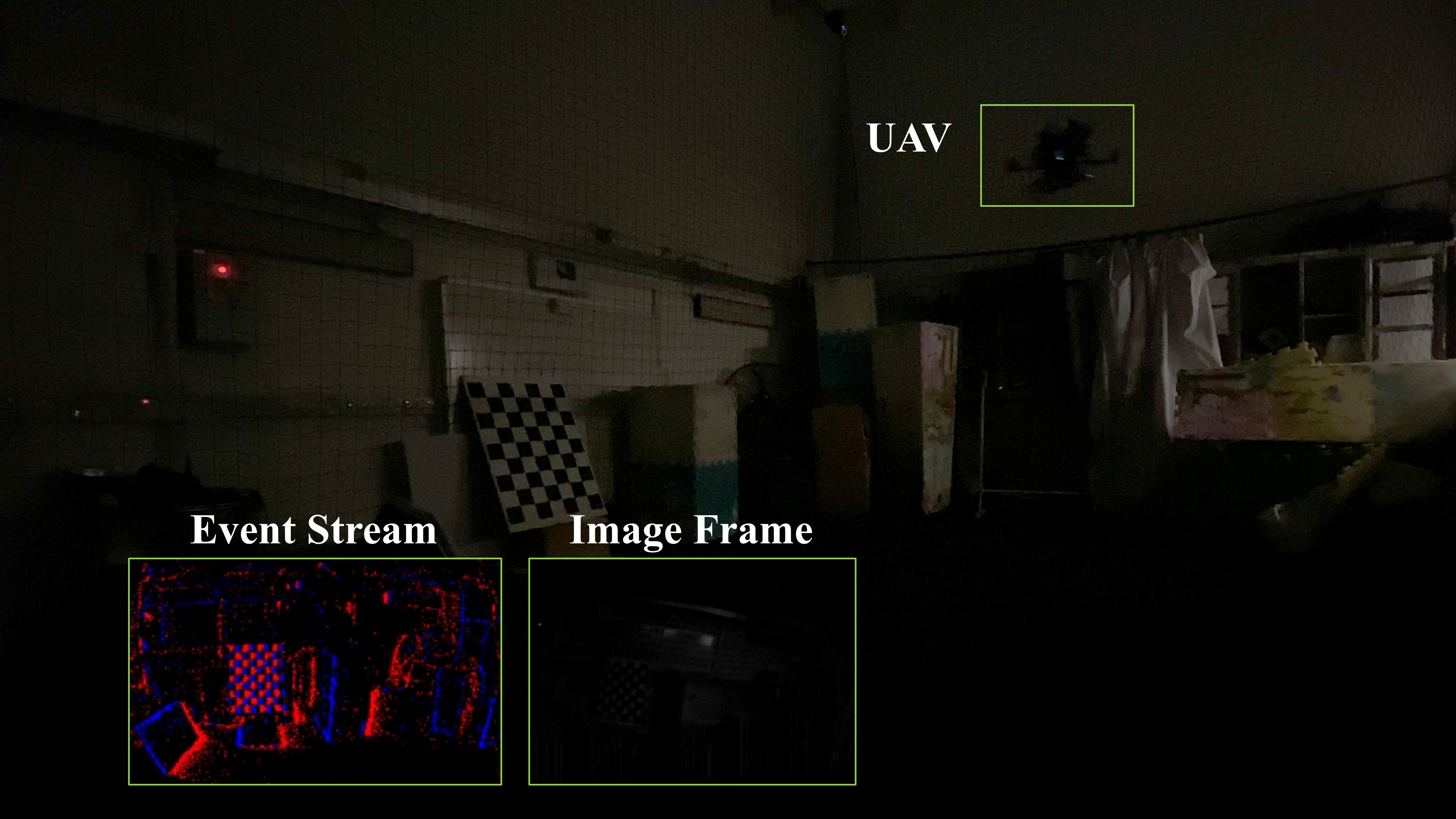}
                \label{figure:drone_testing:a}
                }
        \end{minipage}
        \hspace{1.0em}
        \begin{minipage}[t]{0.8\columnwidth}
                \centering
                \subfigure[ ]{
                \includegraphics[width=\columnwidth, height=0.7\columnwidth]{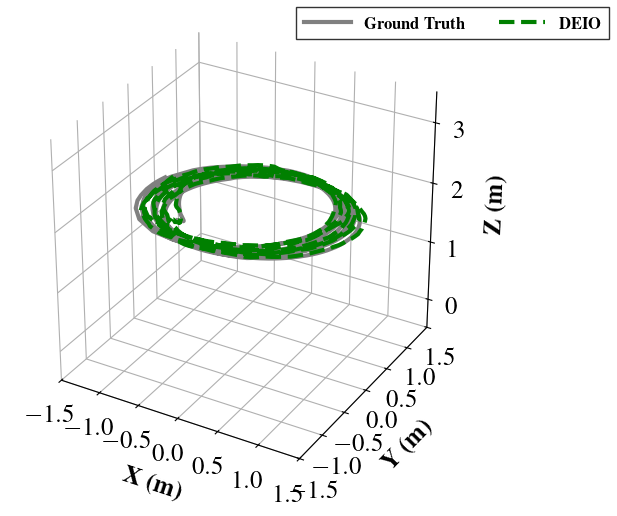}
                \label{figure:drone_testing:b}
                }
        \end{minipage}

        \captionsetup{justification=justified}%图题对齐
        \caption{
        (a) Our quadrotor platform (the image is for visualization only).
        (b) The estimated trajectory of our DEIO in the quadrotor flight and its comparison against the ground truth.
        }  %设置图片的一个编号以及为图片添加标题
        \label{figure:drone_testing}
        % \vspace{-1.6em}%调整表格与正文的距离
\end{figure*}%
%%%%%%%%%%%%%%%%%%***********************************************************************************************************************************************************%%%%%%%%%%%%%%%%%%

%%%%%%%%%%%%%%%%%%%%%%%%%%%%%%%%%%%%%%%%%%%%%%%%%%%%%%%%%%%%%%%%%%%%%%%%%%%%%%%%%%%%%%%%%%%%%%%%%%%%%%%%%%%%
\subsection{Test on Dark Quadrotor-Flight}
\label{section:drone scenarios}
In this section, we evaluate our DEIO in a dark quadrotor flight experiment.
The quadrotor is commanded to track a circle pattern with 1.5 m in radius and 1.8 m in height, shown in Fig.~\ref{figure:drone_testing:b}.
The illuminance in the environment is quite low, resulting in minimal visual information captured by the onboard camera (Fig.~\ref{figure:drone_testing:a}).
The total length of the trajectory is 60.7 m, with an MPE of 0.15 and an average pose tracking error of 9 cm.
% We further illustrate the estimated trajectories (translation and rotation) of our DEIO against the ground truth, as well as their corresponding errors (Fig.~\ref{figure:drone_testing:c}). 
The translation errors in the X, Y, and Z dimensions are all within 0.5 m, while the rotation errors of the Roll and Pitch dimensions are within 6\degree, and the one in the Yaw dimension is within 3\degree.
To the best of our knowledge, this is also the first implementation of monocular event-inertial odometry for pose tracking in dark flight environments, while the previous works~\cite{GWPHKU:Ultimate-SLAM, ESVIO} rely on the image-aided event-IMU estimators.

% %%%%%%%%%%%%%%%%%%%%%%%%%%%%%%%%%%%%%%%%%%%%%%%%%%%%%%%%%%%%%%%%%%%%%%%%%%%%%%%%%%%%%%%%%%%%%%%%%%%%%%%%%%%%
% \subsection{Discussion}
% \label{section:Discussion}

%%%%%%%%%%%%%%%%%%%%%%%%%%%%%%%%%%%%%%%%%%%%%%%%%%%%%%%%%%%%%%%%%%%%%%%%%%%%%%%%%%%%%%%%%%%%%%%%%%%%%%%%%%%%
\subsection{Qualitative Analysis}

In the previous section, we presented a quantitative evaluation of our DEIO framework, showcasing its superior performance and exceptional generalization capabilities across ten challenging datasets. 
To the best of our knowledge, this work constitutes the most comprehensive benchmarking effort in the domain of event-based SLAM to date.
The detailed characteristics of these ten benchmarks, including their inherent challenges and comparative analyses with baseline methods, have been extensively discussed. 
Building on this foundation, this section delves into a qualitative assessment of these demanding scenarios.
Fig.~\ref{figure:deio_eds}(b) depicts the evaluation of DEIO in an environment characterized by extreme darkness (as evident in the image view) and interference from a ceiling-mounted motion capture system. 
Despite these adverse conditions, our framework demonstrated remarkable robustness, with the estimated trajectory closely aligning with the ground truth.
Similarly, Fig.~\ref{figure:deio_FPV} highlights the efficacy of DEIO in a high-speed drone flight scenario. 
This setting introduced additional challenges, including low-texture environments, intense lighting conditions, and motion blur induced by rapid movement. 
Notably, neither learning-based sparse VO (DPVO) nor learning-based dense VIO (DBA-Fusion) can process any sequence successfully under these conditions. 
In contrast, DEIO exhibited both robustness and high precision. 
Further video demonstrations, which provide a more intuitive representation of DEIO, are available on the project website.

% %%%%%%%%%%%%%%%%%%%%%%%%%%%%%%%%%%%%%%%%%%%%%%%%%%%%%%%%%%%%%%%%%%%%%%%%%%%%%%%%%%%%%%%%%%%%%%%%%%%%%%%%%%%%
% \subsubsection{Ablation Study on the Number of Event Patches}

%%%%%%%%%%%%%%%%%%%%%%%%%%%%%%%%%%%%%%%%%%%%%%%%%%%%%%%%%%%%%%%%%%%%%%%%%%%%%%%%%%%%%%%%%%%%%%%%%%%%%%%%%%%%
\subsection{Ablation Study and Runtime Analysis}

In this section, we conduct an ablation study to systematically investigate the impact of the number of event patches per voxel on the system's performance.
The quantity of event patches plays a critical role in the strength of learning-based event data association, as a larger number of patches introduces more event constraints into the factor graph. 
Through carefully designed experiments, we aim to determine the optimal patch configuration that strikes a balance between computational efficiency and system performance.
The results are presented in Table~\ref{table:Ablation_Study_davis240c}, on each sequence, we run five trials and report the median result.
% As observed, increasing the number of event patches per voxel enhances overall performance. 
% However, as shown in Fig.~\ref{figure:real_time}, a higher patch count also reduces the voxel rate of the pose estimation output. 
% To strike an optimal balance between performance and frame rate, we have set 96 patches per voxel as the default configuration in our experiments.
% This analysis yields valuable insights into the interplay between event processing granularity and overall system performance.

%%%%%%%%%%%%%%%%%%***********************************************************************************************************************************************************%%%%%%%%%%%%%%%%%%
\begin{table}[htb]
        \begin{center}
        \captionsetup{justification=justified}
        \caption{
        Accuracy comparison [MPE(\%)] of DEIO with different numbers of event patches in DAVIS240c dataset~\cite{GWPHKU:event-camera-dataset_davis240c}.
        % Unit: MPE(\%).
        }
        \label{table:Ablation_Study_davis240c}
        \resizebox{1.0\columnwidth}{!}
        { 
        \begin{threeparttable}
        \renewcommand{\arraystretch}{1.0}%调整表线和单元格内容之间的距离
        \small  % \large % 增大基础字号
        \begin{tabular}{c|ccc} 
        \hline
        % \makecell{Number of Event-patch \\ Per Frame} & 48 & 96 & 120\\
        \makecell{Event Patches Per Voxel} & 48 & 96 & 120\\
        \hline 
        \makecell{boxes\_translation}    & 0.07 & 0.04  &  0.04 \\
        \makecell{hdr\_boxes}            & 0.09 & 0.08  &  0.09 \\
        \makecell{boxes\_6dof}           & 0.07 & 0.08  &  0.08\\
        \makecell{dynamic\_translation}  & 0.08 & 0.06  &  0.06\\
        \makecell{dynamic\_6dof}         & 0.06 & 0.05  &  0.06\\
        \makecell{poster\_translation}   & 0.04 & 0.08   &  0.04\\
        \makecell{hdr\_poster}           & 0.08 & 0.06  &  0.06\\
        \makecell{poster\_6dof}          & 0.08 & 0.07  &  0.08\\
        \hline   
        Average                          & 0.071 & 0.065  & 0.063  \\
        \hline       
        \end{tabular}
        \end{threeparttable} 
        }
        \end{center}
\end{table}
%%%%%%%%%%%%%%%%%%***********************************************************************************************************************************************************%%%%%%%%%%%%%%%%%%

\begin{figure}[htb]  %%(h 此处（here） t 页顶（top）b 页底（bottom） p 独立一页（page）)
        \centering     
        \captionsetup{justification=justified}%图题对齐
        \includegraphics[width=1.0\columnwidth]{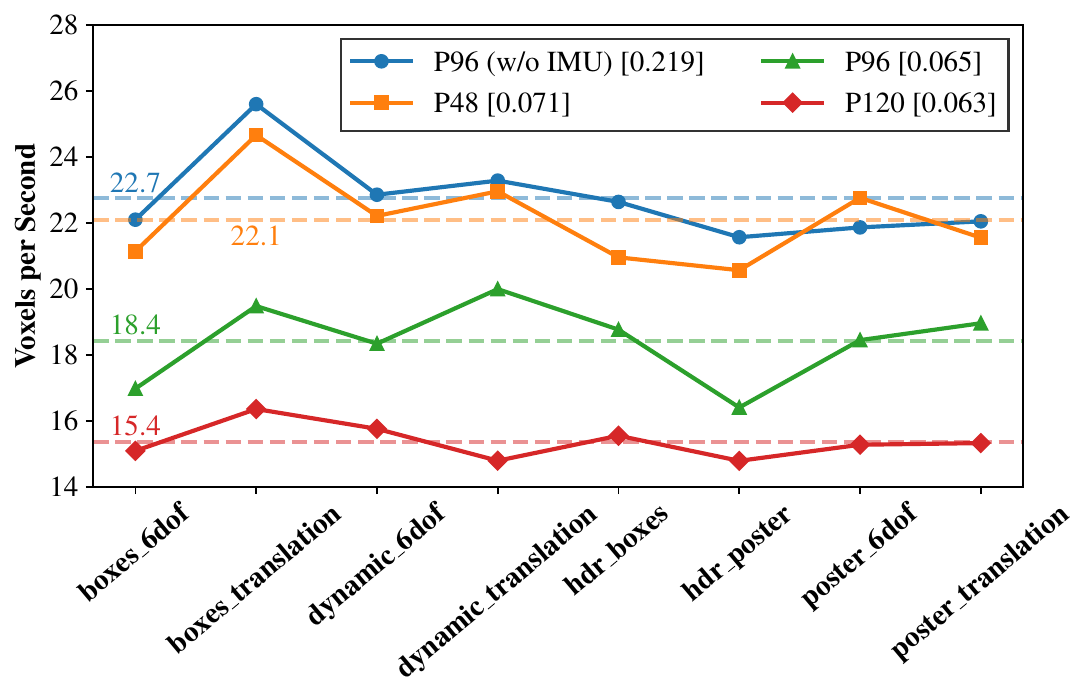}  
        \caption{
        Runtime performance (voxels per second) of our DEIO using 48 (P48), 96 (P96), and 120 (P120) event patches per voxel, as well as a 96-patch version without IMU input (P96 w/o IMU). Values in the brackets indicate the average MPE (\%) over all sequences.
        }  %设置图片的一个编号以及为图片添加标题
        \label{figure:real_time}
        % \vspace{-1.6em}%调整表格与正文的距离
\end{figure}

Additionally, Fig.~\ref{figure:real_time} further illustrates the real-time performance of our DEIO on an Nvidia RTX 3090 GPU.
Our DEIO, configured with 96 patches per event voxel (P96), achieves an average processing speed of 18.4 voxels per second (VPS).
Compared to the variant that excludes IMU data (P96 w/o IMU), the P96 configuration demonstrates a significant accuracy improvement of 69.0\%, with only minimal runtime overhead of 4.3 VPS.
The number of event patches is also a key factor influencing the overall system speed. 
For instance, the P48 variant achieves an average MPE of 0.071 while maintaining a runtime of 22.1 VPS, which is comparable to that of the P96 w/o IMU configuration (22.7 VPS).
However, increasing the number of event patches further (P120) leads to diminishing returns in accuracy improvement while significantly increasing computational demands.
To achieve an optimal balance between performance and frame rate, we have adopted 96 patches per voxel as the default configuration in our experiments.

%%%%%%%%%%%%%%%%%%***********************************************************************************************************************************************************%%%%%%%%%%%%%%%%%%
%%%%%%%%%%%%%%%%%%***********************************************************************************************************************************************************%%%%%%%%%%%%%%%%%%

%%%%%%%%%%%%%%%%%%%%%%%%%%%%%%%%%%%%%%%%%%%%%%%%%%%%%%%%%%%%%%%%%%%%%%%%%%%%%%%%%%%%%%%%%%%%%%%%%%%%%%%%%%%%
\subsection{Discussion}

It is worth noting that this work adopts a learning-based approach exclusively for event data association, while maintaining graph-based optimization for IMU integration. 
This design is driven by two primary considerations: 
(1) learning-based IMU bias estimation methods often exhibit limited generalization across different IMU hardware due to variations in specifications and operating frequencies~\cite{buchanan2022deep}, and 
(2) Such methods rely on labeled training data, which is challenging to obtain, and their computational requirements can compromise real-time performance. 
Our proposed learning-optimization combined framework effectively addresses these challenges, demonstrating remarkable generalization capabilities, as evidenced in the previous section. 
Despite being trained solely on synthetic data, our approach outperforms over 20 state-of-the-art methods across 10 challenging real-world event benchmarks while maintaining real-time efficiency.

%%%%%%%%%%%%%%%%%%%%%%%%%%%%%%%%%%%%%%%%%%%%%%%%%%%%%%%%%%%%%%%%%%%%%%%%%%%%%%%%%%%%%%%%%%%%%%%%%%%%%%%%%%%%
\section{Conclusion}
\label{section:Conclusion}

In this paper, we propose DEIO, a deep learning-based event-inertial odometry method
An event-based deep neural network is utilized to provide accurate and sparse associations of event patches over time, and DEIO further tightly integrates it with the IMU during the graph-based optimization process to provide robust 6 DoF pose tracking.
Evaluation on \textit{ten} challenging event-based benchmarks demonstrates that DEIO outperforms both image-based and event-based baselines.
We have shown that the learning-optimization combined framework for SLAM is a promising direction.
To further enhance the robustness and efficiency of the system, future work will focus on exploring IMU-bias online learning, event-image complementarity, and loop closure mechanisms for learning-based event-SLAM with global BA.

%%%%%%%%%%%%%%%%%%%%%%%%%%%%%%%%%%%%%%%%%%%%%%%%%%%%%%%%%%%%%%%%%%%%%%%%%%%%%%%%%%%%%%%%%%%%%%%%%%%%%%%%%%%%
\bibliographystyle{IEEEtran} % 参考文献排版风格，这个是IEEE transaction的，其他可以自查
\bibliography{references.bib} % 导入bib，references为“references.bib"的文件名

\begin{thebibliography}{10}
\providecommand{\url}[1]{#1}
\csname url@rmstyle\endcsname
\providecommand{\newblock}{\relax}
\providecommand{\bibinfo}[2]{#2}
\providecommand\BIBentrySTDinterwordspacing{\spaceskip=0pt\relax}
\providecommand\BIBentryALTinterwordstretchfactor{4}
\providecommand\BIBentryALTinterwordspacing{\spaceskip=\fontdimen2\font plus
\BIBentryALTinterwordstretchfactor\fontdimen3\font minus \fontdimen4\font\relax}
\providecommand\BIBforeignlanguage[2]{{%
\expandafter\ifx\csname l@#1\endcsname\relax
\typeout{** WARNING: IEEEtran.bst: No hyphenation pattern has been}%
\typeout{** loaded for the language `#1'. Using the pattern for}%
\typeout{** the default language instead.}%
\else
\language=\csname l@#1\endcsname
\fi
#2}}

\bibitem{GWPHKU:EVO}
H.~Rebecq, T.~Horstsch{\"a}fer, G.~Gallego, and D.~Scaramuzza, ``Evo: A geometric approach to event-based 6-dof parallel tracking and mapping in real time,'' \emph{IEEE Robotics and Automation Letters}, vol.~2, no.~2, pp. 593--600, 2016.

\bibitem{GWPHKU:ESVO}
Y.~Zhou, G.~Gallego, and S.~Shen, ``Event-based stereo visual odometry,'' \emph{IEEE Transactions on Robotics}, vol.~37, no.~5, pp. 1433--1450, 2021.

\bibitem{GWPHKU:Ultimate-SLAM}
A.~R. Vidal, H.~Rebecq, T.~Horstschaefer, and D.~Scaramuzza, ``Ultimate slam? combining events, images, and imu for robust visual slam in hdr and high-speed scenarios,'' \emph{IEEE Robotics and Automation Letters}, vol.~3, no.~2, pp. 994--1001, 2018.

\bibitem{EDS}
J.~Hidalgo-Carri{\'o}, G.~Gallego, and D.~Scaramuzza, ``Event-aided direct sparse odometry,'' in \emph{Proceedings of the IEEE/CVF Conference on Computer Vision and Pattern Recognition}, 2022, pp. 5781--5790.

\bibitem{GWPHKU:PL-EVIO}
W.~Guan, P.~Chen, Y.~Xie, and P.~Lu, ``Pl-evio: Robust monocular event-based visual inertial odometry with point and line features,'' \emph{IEEE Transactions on Automation Science and Engineering}, 2023.

\bibitem{zuo2024cross}
Y.-F. Zuo, W.~Xu, X.~Wang, Y.~Wang, and L.~Kneip, ``Cross-modal semi-dense 6-dof tracking of an event camera in challenging conditions,'' \emph{IEEE Transactions on Robotics}, 2024.

\bibitem{zuo2022devo}
Y.-F. Zuo, J.~Yang, J.~Chen, X.~Wang, Y.~Wang, and L.~Kneip, ``Devo: Depth-event camera visual odometry in challenging conditions,'' in \emph{2022 International Conference on Robotics and Automation (ICRA)}.\hskip 1em plus 0.5em minus 0.4em\relax IEEE, 2022, pp. 2179--2185.

\bibitem{FAST-LIEO}
Z.~Wang, Y.~Ge, K.~Dong, I.-M. Chen, and J.~Wu, ``Fast-lieo: Fast and real-time lidar-inertial-event-visual odometry,'' \emph{IEEE Robotics and Automation Letters}, 2024.

\bibitem{GWPHKU:ECMD}
P.~Chen, W.~Guan, F.~Huang, Y.~Zhong, W.~Wen, L.-T. Hsu, and P.~Lu, ``Ecmd: An event-centric multisensory driving dataset for slam,'' \emph{IEEE Transactions on Intelligent Vehicles}, 2023.

\bibitem{DEVO}
S.~Klenk, M.~Motzet, L.~Koestler, and D.~Cremers, ``Deep event visual odometry,'' in \emph{2024 International Conference on 3D Vision (3DV)}.\hskip 1em plus 0.5em minus 0.4em\relax IEEE, 2024, pp. 739--749.

\bibitem{rampvo}
R.~Pellerito, M.~Cannici, D.~Gehrig, J.~Belhadj, O.~Dubois-Matra, M.~Casasco, and D.~Scaramuzza, ``Deep visual odometry with events and frames,'' in \emph{IEEE/RSJ International Conference on Intelligent Robots (IROS)}, June 2024.

\bibitem{DPVO}
Z.~Teed, L.~Lipson, and J.~Deng, ``Deep patch visual odometry,'' \emph{Advances in Neural Information Processing Systems}, vol.~36, 2024.

\bibitem{GWPHKU:kim2016real}
H.~Kim, S.~Leutenegger, and A.~J. Davison, ``Real-time 3d reconstruction and 6-dof tracking with an event camera,'' in \emph{European Conference on Computer Vision}.\hskip 1em plus 0.5em minus 0.4em\relax Springer, 2016, pp. 349--364.

\bibitem{GWPHKU:EMVS}
H.~Rebecq, G.~Gallego, E.~Mueggler, and D.~Scaramuzza, ``Emvs: Event-based multi-view stereo—3d reconstruction with an event camera in real-time,'' \emph{International Journal of Computer Vision}, vol. 126, no.~12, pp. 1394--1414, 2018.

\bibitem{gallego2017event}
G.~Gallego, J.~E. Lund, E.~Mueggler, H.~Rebecq, T.~Delbruck, and D.~Scaramuzza, ``Event-based, 6-dof camera tracking from photometric depth maps,'' \emph{IEEE transactions on pattern analysis and machine intelligence}, vol.~40, no.~10, pp. 2402--2412, 2017.

\bibitem{CMax-SLAM}
S.~Guo and G.~Gallego, ``Cmax-slam: Event-based rotational-motion bundle adjustment and slam system using contrast maximization,'' \emph{IEEE Transactions on Robotics}, 2024.

\bibitem{GWPHKU:Event-based-visual-inertial-odometry}
A.~Zihao~Zhu, N.~Atanasov, and K.~Daniilidis, ``Event-based visual inertial odometry,'' in \emph{Proceedings of the IEEE Conference on Computer Vision and Pattern Recognition}, 2017, pp. 5391--5399.

\bibitem{GWPHKU:ETH-EVIO}
H.~Rebecq, T.~Horstschaefer, and D.~Scaramuzza, ``Real-time visual-inertial odometry for event cameras using keyframe-based nonlinear optimization,'' in \emph{British Machine Vision Conference (BMVC)}, 2017.

\bibitem{GWPHKU:MyEVIO}
W.~Guan and P.~Lu, ``Monocular event visual inertial odometry based on event-corner using sliding windows graph-based optimization,'' in \emph{2022 IEEE/RSJ International Conference on Intelligent Robots and Systems (IROS)}.\hskip 1em plus 0.5em minus 0.4em\relax IEEE, 2022, pp. 2438--2445.

\bibitem{chamorro2023event}
W.~Chamorro, J.~Sol{\`a}, and J.~Andrade-Cetto, ``Event-imu fusion strategies for faster-than-imu estimation throughput,'' in \emph{Proceedings of the IEEE/CVF Conference on Computer Vision and Pattern Recognition}, 2023, pp. 3975--3982.

\bibitem{EKLT-VIO}
F.~Mahlknecht, D.~Gehrig, J.~Nash, F.~M. Rockenbauer, B.~Morrell, J.~Delaune, and D.~Scaramuzza, ``Exploring event camera-based odometry for planetary robots,'' \emph{IEEE Robotics and Automation Letters (RA-L)}, 2022.

\bibitem{GWPHKU:EKLT}
D.~Gehrig, H.~Rebecq, G.~Gallego, and D.~Scaramuzza, ``Eklt: Asynchronous photometric feature tracking using events and frames,'' \emph{International Journal of Computer Vision}, vol. 128, no.~3, pp. 601--618, 2020.

\bibitem{ESVIO}
P.~Chen, W.~Guan, and P.~Lu, ``Esvio: Event-based stereo visual inertial odometry,'' \emph{IEEE Robotics and Automation Letters}, vol.~8, no.~6, pp. 3661--3668, 2023.

\bibitem{ESVO_IMU}
J.~Niu, S.~Zhong, and Y.~Zhou, ``Imu-aided event-based stereo visual odometry,'' pp. 11\,977--11\,983, 2024.

\bibitem{GWPHKU:EVI-SAM}
W.~Guan, P.~Chen, H.~Zhao, Y.~Wang, and P.~Lu, ``Evi-sam: Robust, real-time, tightly-coupled event--visual--inertial state estimation and 3d dense mapping,'' \emph{Advanced Intelligent Systems}, p. 2400243, 2024.

\bibitem{zhu2019unsupervised}
A.~Z. Zhu, L.~Yuan, K.~Chaney, and K.~Daniilidis, ``Unsupervised event-based learning of optical flow, depth, and egomotion,'' in \emph{Proceedings of the IEEE/CVF Conference on Computer Vision and Pattern Recognition}, 2019, pp. 989--997.

\bibitem{hidalgo2020learning}
J.~Hidalgo-Carri{\'o}, D.~Gehrig, and D.~Scaramuzza, ``Learning monocular dense depth from events,'' in \emph{2020 International Conference on 3D Vision (3DV)}.\hskip 1em plus 0.5em minus 0.4em\relax IEEE, 2020, pp. 534--542.

\bibitem{gehrig2021combining}
D.~Gehrig, M.~R{\"u}egg, M.~Gehrig, J.~Hidalgo-Carri{\'o}, and D.~Scaramuzza, ``Combining events and frames using recurrent asynchronous multimodal networks for monocular depth prediction,'' \emph{IEEE Robotics and Automation Letters}, vol.~6, no.~2, pp. 2822--2829, 2021.

\bibitem{ahmed2021deep}
S.~H. Ahmed, H.~W. Jang, S.~N. Uddin, and Y.~J. Jung, ``Deep event stereo leveraged by event-to-image translation,'' in \emph{Proceedings of the AAAI Conference on Artificial Intelligence}, vol.~35, no.~2, 2021, pp. 882--890.

\bibitem{DH-PTAM}
A.~Soliman, F.~Bonardi, D.~Sidib{\'e}, and S.~Bouchafa, ``Dh-ptam: a deep hybrid stereo events-frames parallel tracking and mapping system,'' \emph{IEEE Transactions on Intelligent Vehicles}, 2024.

\bibitem{e-RAFT}
M.~Gehrig, M.~Millh{\"a}usler, D.~Gehrig, and D.~Scaramuzza, ``E-raft: Dense optical flow from event cameras,'' in \emph{2021 International Conference on 3D Vision (3DV)}.\hskip 1em plus 0.5em minus 0.4em\relax IEEE, 2021, pp. 197--206.

\bibitem{RAFT}
Z.~Teed and J.~Deng, ``Raft: Recurrent all-pairs field transforms for optical flow,'' in \emph{Computer Vision--ECCV 2020: 16th European Conference, Glasgow, UK, August 23--28, 2020, Proceedings, Part II 16}.\hskip 1em plus 0.5em minus 0.4em\relax Springer, 2020, pp. 402--419.

\bibitem{GWPHKU:VINS-MONO}
T.~Qin, P.~Li, and S.~Shen, ``Vins-mono: A robust and versatile monocular visual-inertial state estimator,'' \emph{IEEE Transactions on Robotics}, vol.~34, no.~4, pp. 1004--1020, 2018.

\bibitem{GWPHKU:VINS-MONO-initialization}
T.~Qin and S.~Shen, ``Robust initialization of monocular visual-inertial estimation on aerial robots,'' in \emph{2017 IEEE/RSJ International Conference on Intelligent Robots and Systems (IROS)}.\hskip 1em plus 0.5em minus 0.4em\relax IEEE, 2017, pp. 4225--4232.

\bibitem{Droid-slam}
Z.~Teed and J.~Deng, ``Droid-slam: Deep visual slam for monocular, stereo, and rgb-d cameras,'' \emph{Advances in neural information processing systems}, vol.~34, pp. 16\,558--16\,569, 2021.

\bibitem{Tartanair}
W.~Wang, D.~Zhu, X.~Wang, Y.~Hu, Y.~Qiu, C.~Wang, Y.~Hu, A.~Kapoor, and S.~Scherer, ``Tartanair: A dataset to push the limits of visual slam,'' in \emph{2020 IEEE/RSJ International Conference on Intelligent Robots and Systems (IROS)}.\hskip 1em plus 0.5em minus 0.4em\relax IEEE, 2020, pp. 4909--4916.

\bibitem{rebecq2018esim}
H.~Rebecq, D.~Gehrig, and D.~Scaramuzza, ``Esim: an open event camera simulator,'' in \emph{Conference on robot learning}.\hskip 1em plus 0.5em minus 0.4em\relax PMLR, 2018, pp. 969--982.

\bibitem{GWPHKU:event-camera-dataset_davis240c}
E.~Mueggler, H.~Rebecq, G.~Gallego, T.~Delbruck, and D.~Scaramuzza, ``The event-camera dataset and simulator: Event-based data for pose estimation, visual odometry, and slam,'' \emph{The International Journal of Robotics Research}, vol.~36, no.~2, pp. 142--149, 2017.

\bibitem{jung2020constrained}
J.~H. Jung and C.~G. Park, ``Constrained filtering-based fusion of images, events, and inertial measurements for pose estimation,'' in \emph{2020 IEEE International Conference on Robotics and Automation (ICRA)}.\hskip 1em plus 0.5em minus 0.4em\relax IEEE, 2020, pp. 644--650.

\bibitem{HASTE-VIO}
I.~Alzugaray and M.~Chli, ``Asynchronous multi-hypothesis tracking of features with event cameras,'' in \emph{2019 International Conference on 3D Vision (3DV)}.\hskip 1em plus 0.5em minus 0.4em\relax IEEE, 2019, pp. 269--278.

\bibitem{IROS2022_EVIO}
B.~Dai, C.~Le~Gentil, and T.~Vidal-Calleja, ``A tightly-coupled event-inertial odometry using exponential decay and linear preintegrated measurements,'' in \emph{2022 IEEE/RSJ International Conference on Intelligent Robots and Systems (IROS)}.\hskip 1em plus 0.5em minus 0.4em\relax IEEE, 2022, pp. 9475--9482.

\bibitem{tang2024monocular}
K.~Tang, X.~Lang, Y.~Ma, Y.~Huang, L.~Li, Y.~Liu, and J.~Lv, ``Monocular event-inertial odometry with adaptive decay-based time surface and polarity-aware tracking,'' in \emph{IEEE/RSJ International Conference on Intelligent Robots (IROS)}, June 2024.

\bibitem{lee2023event}
M.~S. Lee, J.~H. Jung, Y.~J. Kim, and C.~G. Park, ``Event-and frame-based visual-inertial odometry with adaptive filtering based on 8-dof warping uncertainty,'' \emph{IEEE Robotics and Automation Letters}, 2023.

\bibitem{DBA-Fusion}
Y.~Zhou, X.~Li, S.~Li, X.~Wang, S.~Feng, and Y.~Tan, ``Dba-fusion: Tightly integrating deep dense visual bundle adjustment with multiple sensors for large-scale localization and mapping,'' \emph{IEEE Robotics and Automation Letters}, 2024.

\bibitem{TUM-VIE}
S.~Klenk, J.~Chui, N.~Demmel, and D.~Cremers, ``Tum-vie: The tum stereo visual-inertial event dataset,'' in \emph{2021 IEEE/RSJ International Conference on Intelligent Robots and Systems (IROS)}.\hskip 1em plus 0.5em minus 0.4em\relax IEEE, 2021, pp. 8601--8608.

\bibitem{ORB-SLAM3}
C.~Campos, R.~Elvira, J.~J.~G. Rodr{\'\i}guez, J.~M. Montiel, and J.~D. Tard{\'o}s, ``Orb-slam3: An accurate open-source library for visual, visual--inertial, and multimap slam,'' \emph{IEEE Transactions on Robotics}, vol.~37, no.~6, pp. 1874--1890, 2021.

\bibitem{GWPHKU:VINS-Fusion}
T.~Qin, J.~Pan, S.~Cao, and S.~Shen, ``A general optimization-based framework for local odometry estimation with multiple sensors,'' \emph{arXiv preprint arXiv:1901.03638}, 2019.

\bibitem{envio}
J.~H. Jung, Y.~Choe, and C.~G. Park, ``Photometric visual-inertial navigation with uncertainty-aware ensembles,'' \emph{IEEE Transactions on Robotics}, vol.~38, no.~4, pp. 2039--2052, 2022.

\bibitem{MSOC-S-IKF}
Z.~Zhang, Y.~Song, S.~Huang, R.~Xiong, and Y.~Wang, ``Toward consistent and efficient map-based visual-inertial localization: Theory framework and filter design,'' \emph{IEEE Transactions on Robotics}, vol.~39, no.~4, pp. 2892--2911, 2023.

\bibitem{ESVO2}
J.~Niu, S.~Zhong, X.~Lu, S.~Shen, G.~Gallego, and Y.~Zhou, ``Esvo2: Direct visual-inertial odometry with stereo event cameras,'' \emph{IEEE Transactions on Robotics}, 2025.

\bibitem{GWPHKU:evo_package}
M.~Grupp, ``evo: Python package for the evaluation of odometry and slam,'' \emph{Note: https://github. com/MichaelGrupp/evo Cited by: Table}, vol.~7, 2017.

\bibitem{GWPHKU:VECtor}
L.~Gao, Y.~Liang, J.~Yang, S.~Wu, C.~Wang, J.~Chen, and L.~Kneip, ``Vector: A versatile event-centric benchmark for multi-sensor slam,'' \emph{IEEE Robotics and Automation Letters}, 2022.

\bibitem{ES-PTAM}
S.~Ghosh, V.~Cavinato, and G.~Gallego, ``{ES-PTAM}: Event-based stereo parallel tracking and mapping,'' in \emph{European Conference on Computer Vision (ECCV) Workshops}, 2024.

\bibitem{GWPHKU:MVSEC}
A.~Z. Zhu, D.~Thakur, T.~{\"O}zaslan, B.~Pfrommer, V.~Kumar, and K.~Daniilidis, ``The multivehicle stereo event camera dataset: An event camera dataset for 3d perception,'' \emph{IEEE Robotics and Automation Letters}, vol.~3, no.~3, pp. 2032--2039, 2018.

\bibitem{GWPHKU:FPV}
J.~Delmerico, T.~Cieslewski, H.~Rebecq, M.~Faessler, and D.~Scaramuzza, ``Are we ready for autonomous drone racing? the uzh-fpv drone racing dataset,'' in \emph{2019 International Conference on Robotics and Automation (ICRA)}.\hskip 1em plus 0.5em minus 0.4em\relax IEEE, 2019, pp. 6713--6719.

\bibitem{GWPHKU:DSEC}
M.~Gehrig, W.~Aarents, D.~Gehrig, and D.~Scaramuzza, ``Dsec: A stereo event camera dataset for driving scenarios,'' \emph{IEEE Robotics and Automation Letters}, vol.~6, no.~3, pp. 4947--4954, 2021.

\bibitem{buchanan2022deep}
R.~Buchanan, V.~Agrawal, M.~Camurri, F.~Dellaert, and M.~Fallon, ``Deep imu bias inference for robust visual-inertial odometry with factor graphs,'' \emph{IEEE Robotics and Automation Letters}, vol.~8, no.~1, pp. 41--48, 2022.

\end{thebibliography}

% %%%%%%%%%%%%%%%%%%%%%%%%%%%%%%%%%%%%%%%%%%%%%%%%%%%%%%%%%%%%%%%%%%%%%%%%%%%%%%%%%%%%%%%%%%%%%%%%%%%%%%%%%%%%
\vfill %作用是「很平均的. 把水平(或垂直) 空間塞進去」, 能塞多少就塞多少  对应还有\hfill 
\end{document}